\documentclass{article}

\usepackage{amsmath}        
\usepackage{amssymb}        
\usepackage{amsfonts}       
\usepackage{mathtools}      

\usepackage{graphicx}       
\usepackage{caption}        
\usepackage{subcaption}     

\usepackage{booktabs}       
\usepackage{multirow}       
\usepackage{colortbl}       

\usepackage{bm}             
\usepackage{mathrsfs}       

\usepackage[dvipsnames]{xcolor} 
\usepackage[colorlinks=true, linkcolor=blue, citecolor=blue, urlcolor=blue]{hyperref}  

\usepackage{algorithm}         
\usepackage{algpseudocode}   

\usepackage{siunitx}           
\usepackage{url}               

\usepackage{tcolorbox}         

\usepackage[preprint]{corl_2025} 

\title{GraspMolmo: Generalizable Task-Oriented Grasping via Large-Scale Synthetic Data Generation}

\author{
    Abhay Deshpande$^1$, Yuquan Deng$^1$, Arijit Ray$^2$, Jordi Salvador$^1$, Winson Han$^1$,\\  
    \textbf{Jiafei Duan$^{1,3}$, Kuo-Hao Zeng$^1$, Yuke Zhu$^4$, Ranjay Krishna$^{1,3}$, Rose Hendrix$^1$} \\
    $^1$PRIOR @ Allen Institute for AI, $^2$Boston University, \\
    $^3$University of Washington, $^4$University of Texas at Austin \\
    Corresponding Author: \texttt{abhayd@allenai.org}
}

\newcommand{\ours}{\text{GraspMolmo}}

\newcommand{\ourdata}{\text{PRISM}}
\newcommand{\ourtrainingdata}{\text{\ourdata-Train}}
\newcommand{\ourtestdata}{\text{\ourdata-Test}}
\newcommand{\ourrealdata}{\text{\ourdata-Real}}

\newcommand{\prismfull}{\underline{P}urpose-driven 
    \underline{R}obotic 
    \underline{I}nteraction in 
    \underline{S}cene 
    \underline{M}anipulation}

\makeatletter
\newcommand{\verbatimfont}[1]{\def\verbatim@font{#1}}%

\begin{document}
\maketitle


\begin{abstract}
We present \ours{}, a generalizable open-vocabulary task-oriented grasping (TOG) model. \ours{} predicts semantically appropriate, stable grasps conditioned on a natural language instruction and a single RGB-D frame. For instance, given ``pour me some tea,'' \ours{} selects a grasp on a teapot handle rather than its body. Unlike prior TOG methods, which are limited by small datasets, simplistic language, and unrealistically simple scenes, \ours{} learns from \ourdata{}, a novel large-scale synthetic dataset of 379k samples featuring complex environments and diverse, realistic task descriptions. We fine-tune the Molmo vision-language model on this data, enabling \ours{} to generalize to novel open-vocabulary instructions and objects. In challenging real-world evaluations, \ours{} achieves state-of-the-art results, with a 70\% prediction success on complex tasks, compared to the 35\% achieved by the next best alternative. 
\ours{} also successfully demonstrates the ability to predict semantically correct bimanual grasps zero-shot.
We release our synthetic dataset, code, model, and benchmarks to accelerate research in task-semantic robotic manipulation, which, along with videos, are available at \href{https://abhaybd.github.io/GraspMolmo/}{this~URL}.

\end{abstract}


\begin{figure}[ht]
    \centering
    \includegraphics[width=0.99\textwidth]{media/Teaser_6.pdf}
    \caption{\textbf{A.} We introduce a large dataset of task-semantic-annotated tasks in virtual scenes with a large variety of target objects, which we name \prismfull{}. \textbf{B.} We use \ourdata{} to train \ours{}, which represents grasps as points and matches them to grasp proposals. \textbf{C.} We show strong transfer and generalization in real-world complex scenes with interesting task semantics using a 7-DoF Franka FR3 arm, and additionally show zero-shot adaptation to semantically-appropriate bimanual grasping.}
    \label{fig:teaser}
\end{figure}

\section{Introduction}

Robotic grasping has progressed significantly, with models now capable of robustly predicting stable grasps for a wide range of objects. Yet most existing methods operate in an object-centric fashion: they predict grasps that are stable for a given object but oblivious to the task at hand. In the real world, however, how an object should be grasped depends on what the robot is trying to do with it.

Consider a kitchen knife. A task-agnostic grasping system might simply choose a stable grasp anywhere on the knife, including the blade.
While the blade might be an appropriate grasp site when handing the knife safely to another person, it would be ineffective for the purposes of chopping vegetables.
These task-specific differences highlight the need for task-oriented grasping (TOG): The same physical object requires dramatically different grasps depending on the intended task \cite{detry2017task, murali2020taskgrasp}.

Recent work has introduced relevant datasets and benchmarks~\cite{murali2020taskgrasp,fang2018tog,tang2023graspclip}, but they are limited in realism and diversity.
Some contain relatively simple scenes without any distractors and lack diverse object categories~\cite{murali2020taskgrasp}. Others often templatize (e.g., ``grasp the mug to pour'' or ``grasp the pan to clean''). These simplifications fail to capture the complexity of real-world environments and the wide variety of language humans use to specify tasks, such as ``mince some garlic'' or ``do the dishes''.
Furthermore, many models require pre-segmented point clouds~\cite{tang2023graspclip} or multi-view observations~\cite{fang2018tog}, restricting real-time deployment.

To address these limitations, we propose \ours{}: a generalizable task-conditioned grasping model trained entirely on synthetic data. To overcome the brittleness of prior datasets, we create \prismfull{} (\ourdata{}), a large-scale suite of task-grasp pairs with 10,000 scenes, 2,356 object instances, and task descriptions ranging from simple (``cut the apple'') to compositional or nuanced (``mince some garlic for a salad''). 
This dataset spans diverse, busy scenes with realistic textures and occlusions, and multiple objects per task.

Using this dataset, we fine-tune Molmo~\cite{deitke2024molmopixmoopenweights},
a recent open-weight vision-language model (VLM), to create \ours{}, enabling it to predict 6-DoF grasps from a single RGB-D frame and a task instruction.
\ours{} achieves state-of-the-art results on the TaskGrasp benchmark (76.7\% versus the next best prior work at 72.3\%). \ours{} also achieves significant performance boost on our benchmark of simulated realistic scenes (62.5\% versus 40.0\%). 
We further demonstrate \ours{} transfers zero-shot to the real world in representative household scenes, including objects and tasks completely unseen in training (70.4\% prediction and 61.1\% overall success rates).
\ours{} also zero-shot transfers to semantically-appropriate bimanual grasping in qualitative testing. We will release our dataset, model, code, and benchmarks to enable further research in task-semantic manipulation.

\section{Related work}

\noindent\textbf{Task-oriented grasping.}
The field of Task-Oriented Grasping (TOG) addresses how robots should grasp objects based on intended tasks rather than just stability.
The TaskGrasp dataset~\citep{murali2020taskgrasp} provides training data and evaluation benchmarks for much of the existing TOG literature, but features only simple scenes with limited task diversity and simplistic specification.
Leveraging TaskGrasp or otherwise, recent approaches use diverse strategies with specific limitations.
GraspGPT \cite{Tang2023GraspGPTLSA} and FoundationGrasp \cite{Tang2024FoundationGraspGTA} are most similar to our method, but due to their reliance on TaskGrasp, have limited pretraining with simple scenes and tasks. LERF-TOGO \citep{Kerr2023LanguageERA} uses language-embedded radiance fields to localize grasp points in a scene, but has costly requirements such as multiple views and long inference times. RTAGrasp \citep{dong2024rtagrasp} transfers grasping behavior from human demonstrations but depends heavily on task-grounded data. CROG~\cite{tziafas2023language} and GraspCLIP \citep{Tang2023TaskOrientedGPA} directly infer per-pixel predictions for grasp poses, but are limited to 4-DoF grasps, and strictly require a top-down camera view. TOGNet~\citep{xie2024target} has limited task-aware grounding, resulting in coarse semantics that are ill-suited for diverse tasks. Our work addresses these collective limitations through synthetic data generation with complex scenes, diverse and natural task specifications, and large data scale.

\noindent\textbf{Vision-language models in robotics} 
Recent work has shown that leveraging foundation vision-language models (VLMs) can yield significant progress towards generalizability in robotics~\cite{liu2023llava, deitke2024molmopixmoopenweights, achiam2023gpt, reid2024gemini, zitkovich2023rt2, kim2024openvla, gemini2025robotics, black2024pi0, bjorck2025gr00tn1}. Within this paradigm, some directly use pre-trained VLMs~\cite{liu2024moka, huang2024copa, huang2024rekep, duan2024manipulate}, while others fine-tune these models to enhance specific capabilities, including action prediction~\cite{li2024llara, kim2024openvla, black2024pi0}, physics~\cite{chow2025physbench}, navigation~\cite{Zhang_2024_CVPR, ehsani2024spocimitatingshortestpaths}, and spatial reasoning~\cite{yuan2024robopoint, ray2024sat, song2025robospatialteachingspatialunderstanding}.
Our method aligns with the latter approach, fine-tuning a state-of-the-art VLM~\cite{deitke2024molmopixmoopenweights} specifically for task-semantic grasping tasks. While other works may predict object relationships~\cite{ray2024sat, chow2025physbench} or general pointing~\cite{yuan2024robopoint}, we specifically ground our pointing predictions in object affordances, i.e. how and where an object should be grasped.
In contrast to vision-language-action models (VLAs) that directly predict low-level robot actions~\cite{bjorck2025gr00tn1, black2024pi0, brohan2023rt}, our method directly outputs grasps. This avoids embodiment-specific constraints and shortcuts having to learn redundant behavior, such as free-space motion.
While works like~\cite{ehsani2024spocimitatingshortestpaths, zeng2024poliformerscalingonpolicyrl, yuan2024robopoint, Zhang_2024_CVPR} also predict points or high-level plans that aim to be generalizable, they either only focus on navigation, or don't specifically finetune grasping affordances into the model, resulting in a lack of robustness for grasp prediction.
We also consider recent works~\cite{ray2024sat, silwal2024we, ehsani2024spocimitatingshortestpaths} that show that simulation is a promising and scalable data source for real-world generalization. To that end, we generate large-scale, task-oriented grasping data that focuses on complex task-dependent object manipulation, going beyond simpler QA or navigation tasks in earlier work.

\noindent\textbf{Object affordances in robotics.}
Affordances describe the functional properties of objects, i.e. how they can be manipulated, linking perception to action beyond appearance. Affordance prediction has driven advances in learning-based 6-DoF grasping \cite{sundermeyer2021contact,murali20206,jiang2021synergies} and stable object placement  \cite{zeng2020transporter,liu2022structformer,yuan2023m2t2,yuan2024robopoint}. Researchers have explored various representations, including part-level segmentation \cite{do2018affordancenet}, dense visual descriptors \cite{florencemanuelli2018dense}, and keypoint-based encodings \cite{manuelli2019kpam,qin2020keto,mo2021where2act}. For policies, affordances are learned implicitly from demonstrations and generalized using visual features, as seen in TransporterNet \cite{zeng2020transporter}, CLIPort \cite{shridhar2022cliport}, and recent works leveraging vision-language models (VLMs). However, many of these approaches rely on pretrained VLMs for visual grounding without incorporating deeper semantic reasoning, often resulting in less accurate or unstable point detection. In contrast, our approach fine-tunes Molmo, a VLM enhanced for semantic understanding, to generate task-aware, stable grasp points through contextualized spatial reasoning.

\section{Method}

An overview of our overall method is presented in Fig.~\ref{fig:teaser}. First, we generate a large dataset of synthetic scenes with task-semantic-annotated tasks for each scene object. We then fine-tune a VLM by representing these grasps as points on the image, and then at inference time re-match these points to outputs from a grasp proposal network. An overview of the data generation process for \ourtrainingdata{} and \ourtestdata{} is presented in Fig.~\ref{fig:datagen_overview} and Sec.~\ref{sec:datagen}.

\subsection{Generation of \ourtrainingdata~and \ourtestdata}
\begin{figure}[ht]
    \centering
    \includegraphics[width=0.95\textwidth]{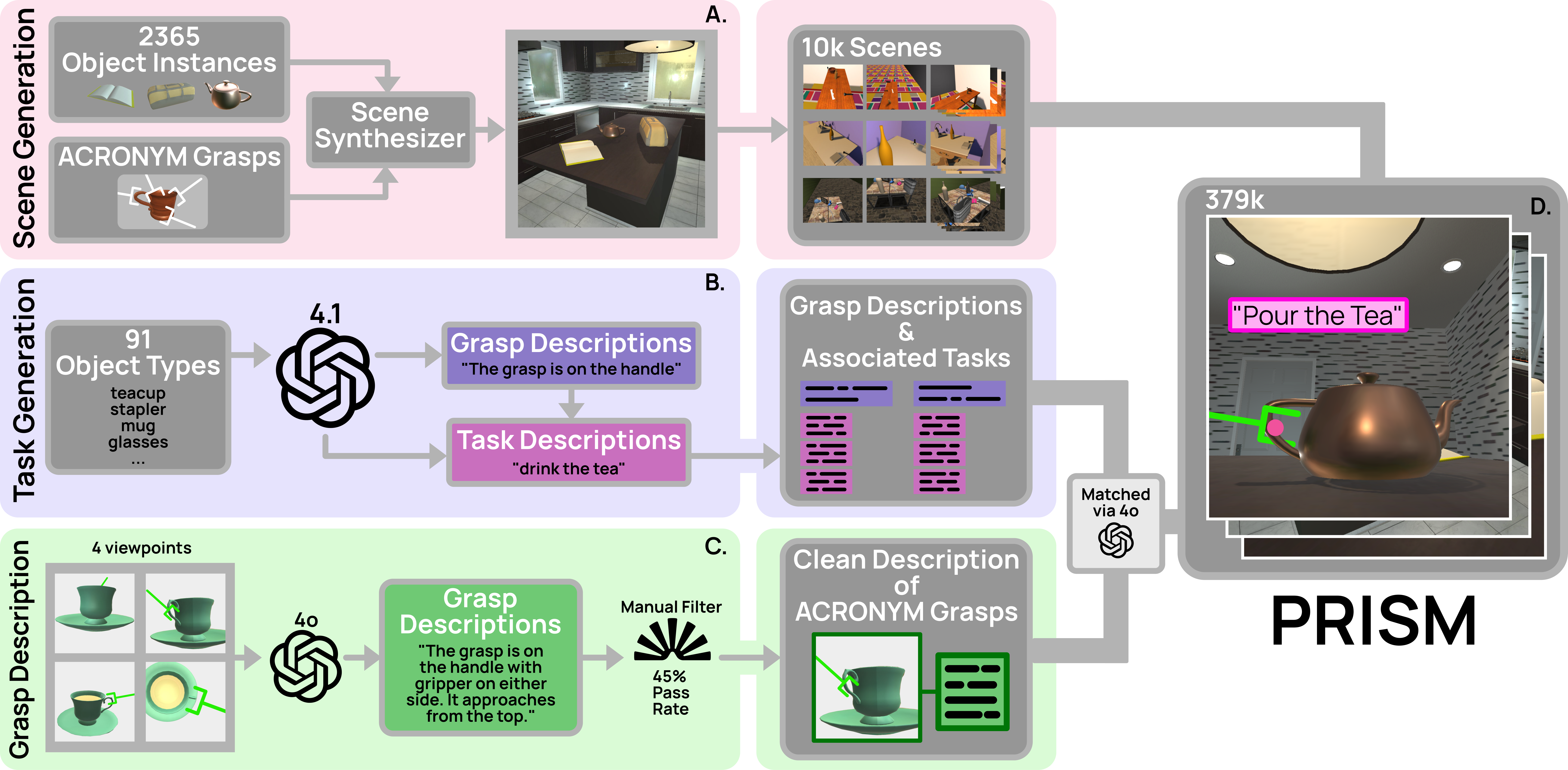}
    \caption{A major contribution is our generated dataset \ourtrainingdata~and evaluation benchmark  \ourtestdata~(Sec.~\ref{sec:datagen}). First, synthetic scenes are generated from Shapenet-Sem \cite{shapenet2015} assets and ACRONYM \cite{acronym2020} grasps. 
    Next, object-centric spatial descriptions of grasps are generated and manually filtered 
    and interesting and varied tasks are generated for object categories.
    }
    \label{fig:datagen_overview}
\end{figure}
\label{sec:datagen}

\paragraph{Scene Generation}
Our dataset employs assets from ShapeNet-Sem \cite{shapenet2015, savva2015semgeo} with grasps derived from ACRONYM \cite{acronym2020}. We enumerate the 91 object classes used in data generation in Table~\ref{tab:object_classes}. These assets are procedurally composed in a scene using SceneSynthesizer~\cite{Eppner2024}, and additional lighting variations and camera randomization are used to increase visual diversity. As illustrated in Figure~\ref{fig:datagen_overview}, this procedure results in a comprehensive collection of scenes spanning different object categories, lighting conditions, and spatial arrangements. Furthermore, we render each scene from 10 distinct camera viewpoints to ensure that our training data captures objects from multiple perspectives. This multi-view approach is particularly important, as grasp affordances often depend on the viewpoint, which can vary wildly in the real world. We track each grasp's visibility from each viewpoint, enabling data filtering to ensure data quality.

\paragraph{Grasp Descriptions as a Bridge} Generating diverse task-grasp annotations at large scale is intractable, as every task-object-grasp triple must be annotated. In other words, annotating $G$ grasps per object for $O$ objects across $T$ tasks is $\mathcal{O}(TOG)$. Our insight is that we can drastically reduce the effort required to generate task-grasp annotations by exploiting common structure. Namely, we note that many different tasks might require grasping an object in the same way. Concretely, pouring from a mug and drinking from it both require grasping the handle in the same manner. Therefore, we use \textit{grasp descriptions} as the bridge between tasks and grasps. Instead of individually annotating each task-object-grasp triple with a binary label, we annotate each object-grasp pair with a natural language description of how the grasp is gripping the object. Separately, for each task we generate descriptions of how an object should be grasped to complete that task. Then, matching tasks to object-grasp pairs reduces to simple text matching. The use of common-sense reasoning and natural language make this task well-suited for automation via LLMs, further aiding in scalability. This decomposition decouples tasks from object-grasps pairs, meaning generating annotations is now $\mathcal{O}(T+OG)$, which is significantly more tractable to perform at large scale.

\paragraph{Grasp Description Generation}

To generate high-quality grasp descriptions, we employ a two-stage approach combining LLMs with human verification. We first use GPT-4o \cite{openai2024gpt4ocard} to generate synthetic grasp descriptions, given multiple viewpoints of an object being grasped. An example of this is illustrated in Fig.~\ref{fig:datagen_overview}. To ensure accuracy and quality, we then engage human workers through the Prolific platform to verify the LLM-generated descriptions, and provide corrections where necessary. We found that 45\% of the synthetically generated grasp descriptions were judged to be accurate while the other 55\% required manual editing, illustrating the importance of human verification for high quality data. Approximately \$3,400 was spent on human data filtering.


\paragraph{Task Generation}
We generate a rich diversity of semantic grasping tasks in two steps, leveraging the knowledge and common-sense reasoning in existing LLMs. Given an object class, we first prompt GPT-4.1 to generate two grasp descriptions that are maximally different but still plausible for manipulation tasks, avoiding hallucinating features or descriptions which may or may not exist for a specific instance of the object class. The LLM then generates four valid semantic tasks for each grasp, while minimizing compatibility of the instruction with the alternative grasp. In total, we generate 728 unique tasks corresponding to two grasps for each of the 91 object classes. Complete prompting details are available in the appendix.

\paragraph{Matching Tasks to Grasps} Finally, we match the generated grasp descriptions for each object-grasp pair with that of the generated tasks, using GPT-4o. For each object in a scene we first determine the visible grasps and corresponding grasp descriptions. For each generated task for that object, we then take the corresponding proposed grasp description and ask GPT-4o to determine which annotated object-grasp pair, if any, describes the same grasp. If the LLM determines that one of the existing grasps in the scene is described by the generated description, the task is paired with the object-grasp to create an task-object-grasp triple, and added to the dataset.

\paragraph{Assembling \ourtrainingdata{} and \ourtestdata}
In total, \ourtrainingdata~ contains 9,424 unique object-centric grasp instances across 2,356 unique objects, which is scaled combinatorially via procedural generation. Although we do not use this information for training \ours{}, the dataset also includes calibrated 3D grasp poses for all objects, providing ground truth spatial information that can be used for both 2D and 3D perception tasks. The final dataset is 378,844 samples: each has a scene render, a task in natural language, a ground-truth semantically appropriate and stable grasp in calibrated camera coordinates, an object-centric spatial description of that grasp in natural language, and a pixel location corresponding to the correct grasp point for that task. We present sample generated scenes from \ourdata{} in Figure~\ref{fig:scenes}, showcasing the diversity in objects, arrangements, and appearances.


\subsection{TaskGrasp-Image}
\label{sec:taskgrasp-image}

TaskGrasp \cite{murali2020taskgrasp} is a standard dataset in the field of task-oriented grasping, used for training and evaluation. The TaskGrasp dataset consists of partial object point clouds, annotated with stable grasp poses. Each grasp on each object is annotated with binary labels, indicating whether each grasp satisfies a given task verb (e.g., scoop, cut, stir). However, the use of point clouds makes using image-based models difficult, especially due to limitations in the data such as noisy point cloud reconstructions caused by self-occlusions and segmentation artifacts. Additionally, the use of single verbs and nouns to define a task results in simplistic task conditioning, which insufficiently captures the richness of real-world human instructions.

To address these challenges, we construct TaskGrasp-Image, a new image-based dataset derived from the existing TaskGrasp dataset. Concretely, TaskGrasp-Image is made up of the original \mbox{RGB-D} images from TaskGrasp instead of the fused point clouds assembled from said images. This yields more realistic input due to the absence of fusion, filtering, or segmentation artifacts. As a result, TaskGrasp-Image preserves the ground-truth grasp annotations while placing them in the context of real, unprocessed RGB-D imagery. We defer complete details on deriving \mbox{TaskGrasp-Image} from TaskGrasp to Appendix~\ref{appendix:taskgrasp-image}.


\subsection{\ours{}}
\label{sec:graspmolmo}

\paragraph{Training} Using \ourtrainingdata~and the training data from TaskGrasp-Image (specifically, split 0 from the task split type), we create \ours{} by fine-tuning Molmo \cite{deitke2024molmopixmoopenweights}, a state-of-the-art VLM for pointing and spatial reasoning tasks, to point to grasps. We co-train on PixMo and all other Molmo training tasks in order to preserve the capability to generalize to unseen objects and settings while adapting to grasping.
We use the natural language object-centric grasp description from our training data as a chain-of-thought processing step, requiring the model to output both the pixel location for the grasp and an object-centric description of that grasp. We sample 45\% and 10\% of our data mixture from \ourtrainingdata~ and TaskGrasp-Image, respectively, and proportionally downweight other data sources. We defer details on data mixture and hyperparameters to Appendix~\ref{appendix:graspmolmo_training_details}, and prompting details to Appendix~\ref{appendix:prompting_details}.

\paragraph{Mapping Output Points to Grasps} \ours{} outputs points on the image plane, which must be matched to a candidate grasp predicted by a stable grasp generator. To do so, we map each grasp to a point on the image, and select the grasp with the closest corresponding point to Molmo's prediction. Concretely, given a set of candidate grasps $\mathcal{G}\subset\text{SE(3)}$, a function $f\colon\mathcal{G}\to\mathbb{R}^2$ that maps a grasp to a pixel coordinate, and the output $p\in\mathbb{R}^2$ from Molmo, we select the grasp $\hat{g}\coloneqq\arg\min_{g\in\mathcal{G}}\|f(g)-p\|$. Additional details on this process are given in Appendix~\ref{appendix:prompting_details}.

\section{Evaluation}
\label{sec:evaluation}
We evaluate \ours{} and baselines on three distinct benchmark settings: a benchmark from literature with simple objects and minimal visual diversity, a synthetic held-out dataset of fully composed scenes with unseen objects, and real-world transfer scenarios. The performance gap between methods widens notably as we progress from simpler to more complex evaluations, revealing fundamental differences in approach capabilities that are not apparent in basic benchmarks.

\begin{table}[H]
    \centering
    \begin{tabular}{lcccc}
        & \multirow{2}{*}{TaskGrasp-Image} & \multirow{2}{*}{\ourtestdata} & \ourrealdata & \ourrealdata \\
        & & & (Prediction) & (Overall) \\
        \cmidrule(r){2-5}
        Random & 54.5\% & 29.3\% & - & - \\
        GraspGPT & 72.3\% & 40.0\% & 35.2\% & 24.1\% \\
        Molmo & 75.6\% & 49.8\% & 35.2\% & 33.3\% \\
        GraspMolmo & \textbf{76.7}\% & \textbf{62.5}\% & \textbf{70.4\%} & \textbf{61.1\%} \\
        \bottomrule
    \end{tabular}
    \smallskip
    \caption{Top-1 accuracy for grasp prediction across increasingly challenging task-oriented grasping settings. For real-world online evaluations, we separately report the prediction success rate (was the predicted grasp correct) and the overall success rate (was the predicted grasp correct \textit{and} did the robot successfully grasp the object).}
    \label{tab:performance}
\end{table}

\subsection{Baselines}
We compare our method to GraspGPT \cite{Tang2023GraspGPTLSA}, the current state-of-the-art approach in Task-Oriented Grasping, with similar assumptions to ours. We use the pretrained \texttt{mode\_t\_split\_0} checkpoint of GraspGPT without modification. We also compare to Molmo \cite{deitke2024molmopixmoopenweights}, which is simply the base VLM we fine-tune to create \ours{}, in order to illustrate the utility of training on \ourdata{}. Finally, we also include a naive random baseline that uniformly samples a grasp from the candidates, underlining the importance of task-oriented reasoning. We follow previous works \cite{murali2020taskgrasp,Tang2023GraspGPTLSA,Tang2024FoundationGraspGTA} in restricting real-world evaluations to single-view RGB-D observations, and in the interest of maximizing similarity with the real-world setting, we do the same for all evaluations.

We note that GraspGPT requires a segmented point cloud of the object, and therefore depends on Segment Anything 2 (SAM2) \cite{ravi2024sam2} and GroundingDINO \cite{liu2023grounding} to extract the object point cloud for downstream processing. Additionally, in addition to the task instruction (e.g. ``give me some water''), GraspGPT also requires access to the specific object being manipulated (e.g. water bottle), and the specific action primitive being performed (e.g. handover). In the real world, such extra information may not be available, limiting its applicability. For GraspGPT in our setting, we use GPT-4o to infer the specific object and task primitive from the task instruction and the image, adding both computational overhead and the potential for error propagation. In contrast to these requirements, \ours{} directly processes raw sensor data and freeform natural language without needing intermediate segmentation or task simplification steps.

\subsection{TaskGrasp-Image}
For the TaskGrasp-Image benchmark, we follow the same data presentation format as detailed in Sec.~\ref{sec:taskgrasp-image} - single-view RGB-D observations rather than fused point clouds. We evaluate on split 0 of the ``task'' split type from TaskGrasp, and illustrate a performance increase over all baselines in Table~\ref{tab:performance}. Following~\cite{Tang2023GraspGPTLSA}, we normalize across tasks for this evaluation.

\subsection{\ourtestdata}
\ourtestdata~is a synthetic evaluation set constructed using the same pipeline as our training data, but with both held-out object instances (of seen classes) and completely novel object classes. We defer full details, including a complete object list and randomization parameters, to Appendix~\ref{appendix:scene_generation_details}. This benchmark tests generalization capabilities across novel objects and novel scenes, while maintaining the controllable diversity and large scale of synthetic data. By evaluating on completely unseen object categories, we demonstrate that \ours{} learns generalizable task-grasp relationships rather than memorizing specific object-grasp pairings.

\begin{figure}[htb]
    \centering
    \begin{subfigure}{0.58\linewidth}
        \centering
        \includegraphics[width=\linewidth]{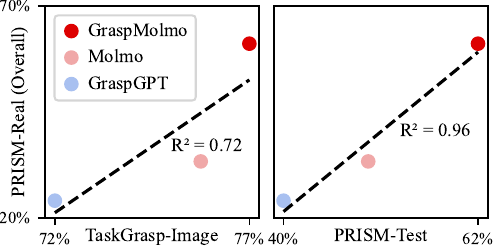}
        \caption{}
        \label{fig:benchmark-correlation}
    \end{subfigure}
    \hfill
    \begin{subfigure}{0.37\linewidth}
        \centering
        \raisebox{0.25cm}{\includegraphics[width=\linewidth]{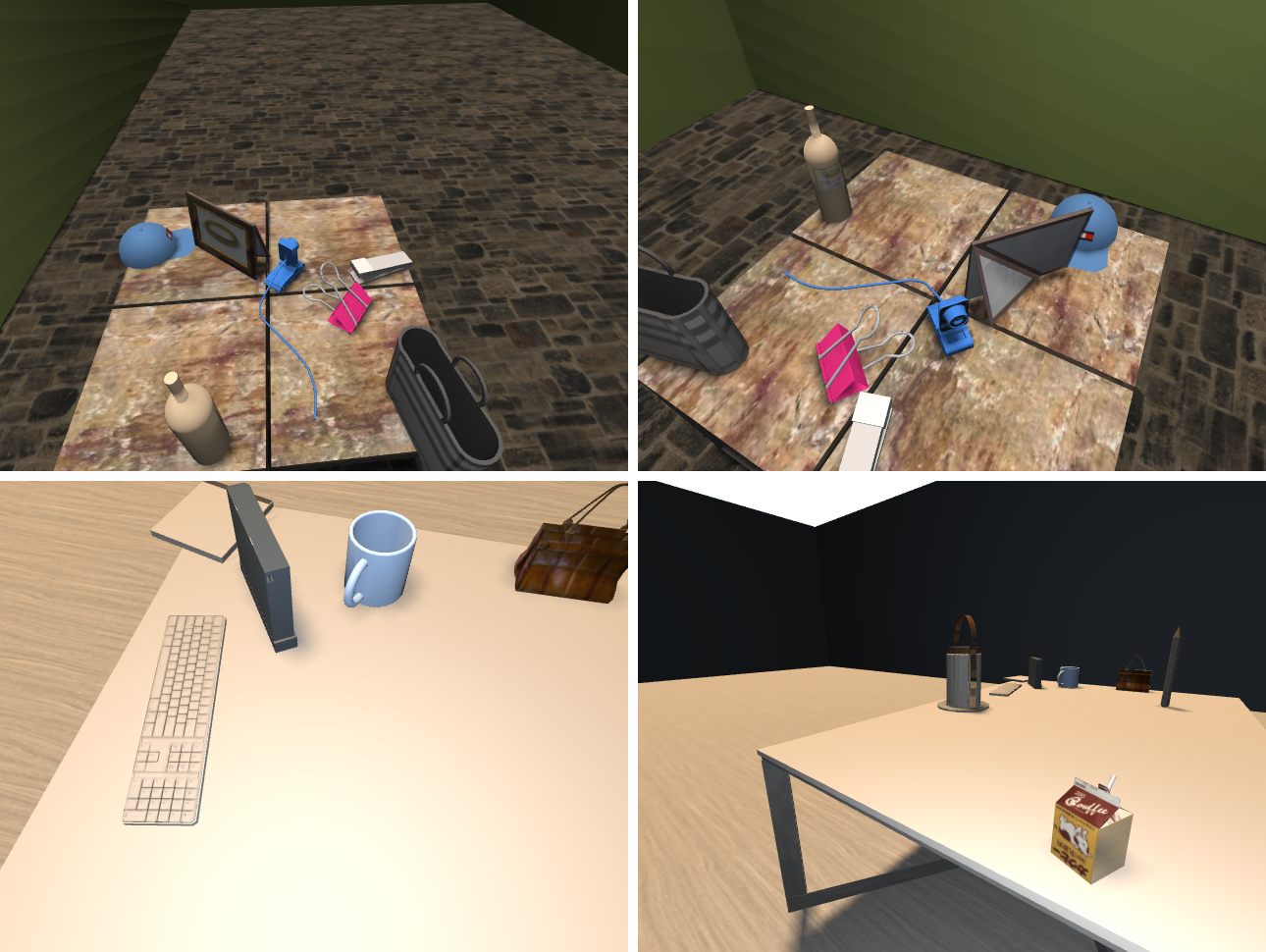}}
        \caption{}
        \label{fig:scenes}
    \end{subfigure}
    \caption{(a) Performance on \ourtestdata{} is a better indicator of success in real-world scenarios than TaskGrasp-Image. (b) Sample scenes and viewpoints, illustrating the object, viewpoint, and lighting diversity of \ourdata{}. }
\end{figure}

In these challenging scenarios with held-out objects and scenes, Table~\ref{tab:performance} shows our method achieves a 62.5\% success rate, while baselines drop below 50\%. This widening performance gap and lower success rates confirm our intuition that the complex scenes and tasks, along with diverse randomization, make \ourtestdata{} a challenging and valuable benchmark. As illustrated in Figure~\ref{fig:benchmark-correlation}, we also see that \ourtestdata{} also correlates well with real-world performance, further proving its utility.

\subsection{Real-World Transfer with \ourrealdata}
\begin{figure}[ht]
    \centering
    \includegraphics[width=0.95\textwidth]{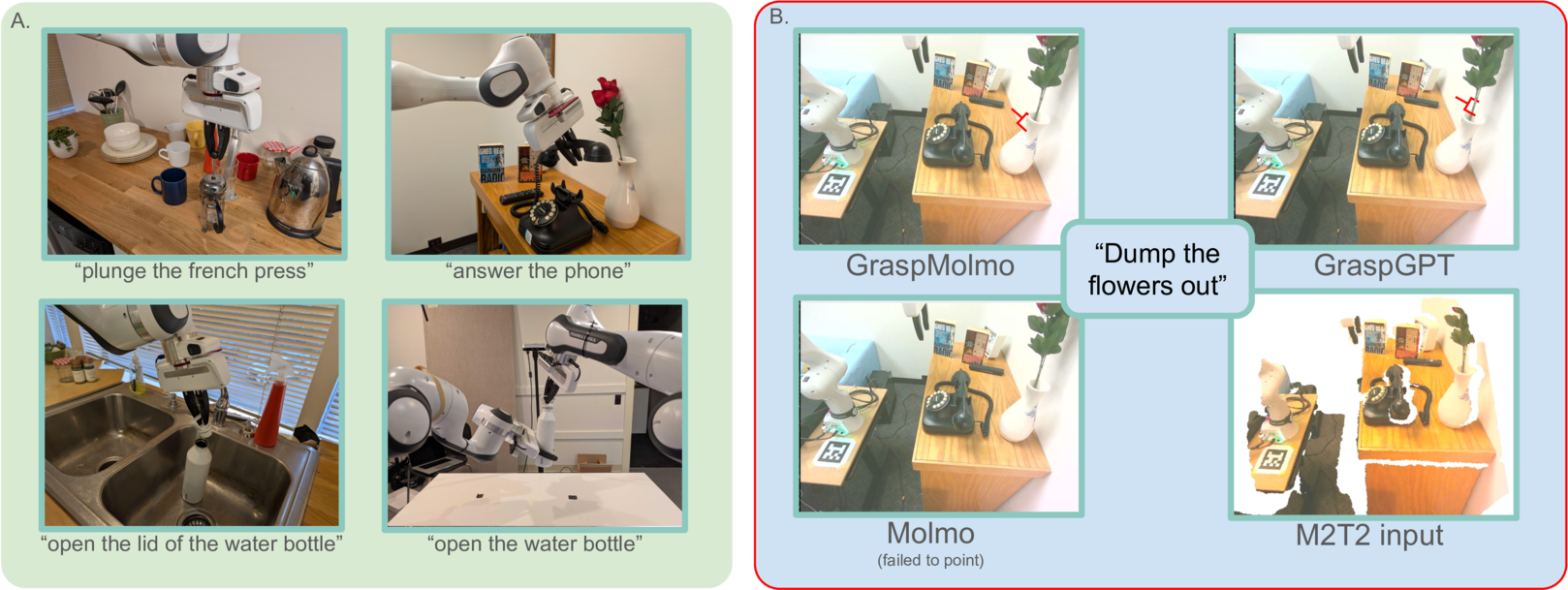}
    \caption{(a) We evaluate on three real-world scenes representative of in-home use cases, with varying objects with diverse task semantics. We also demonstrate zero-shot applicability to bimanual task-oriented grasping. Some task instructions have been shortened for brevity. (b) We illustrate sample grasp outputs from \ours{} and baselines for the task ``dump the flowers out'', where the robot must grasp the vase and turn it over, to empty out the flowers. \ours{} correctly grasps the vase in an optimal position to flip it.}
    \label{fig:qualitative}
\end{figure}

To demonstrate real-world transfer capabilities, we test on a real robot platform. We use M2T2~\cite{yuan2023m2t2} as the stable grasp generator, and execute the predicted grasps on a 7-DoF Franka FR3 arm. We evaluate the considered methods on 3 realistic scenes, featuring a total of 9 household objects representative of in-home applications, each with 2 associated tasks. With 3 trials per task, we perform a total of 54 trials per method. Evaluation scenes and sample tasks are pictured in Fig.~\ref{fig:qualitative}, showcasing the complexity and diversity of our evaluation. For full details about the evaluation scenes, objects, and tasks, please refer to Appendix~\ref{appendix:real_robot_eval_details}.

Not only do our evaluation scenes contain multiple objects with meaningful variation in task semantics, but each object has multiple affordances and associated tasks. This measures a model's ability to reason about how task semantics inform grasping behavior, going beyond simple memorization of task-agnostic affordances. We see in Table~\ref{tab:performance} that \ours{} outperforms baselines by a considerable margin, attaining a 70.4\% accuracy for grasp predictions, double that of the closest baseline. We also report a 61.1\% success rate overall, including execution on the robot, whereas the closest baseline attains 33.3\%. A paired t-test confirms that our improvements over Molmo and GraspGPT are statistically significant at $p<0.05$, validating our real-world evaluation.


\ours{}'s sim-to-real transfer is largely attributable to a few key design choices. Firstly, \ours{} is co-trained with real images, both from Pixmo~\cite{deitke2024molmopixmoopenweights} and real grasping data from TaskGrasp-Image. As shown in Table~\ref{tab:data-mixture}, real images comprise 55\% of training data, helping to combat overfitting to unrealistic simulation renderings. Additionally, PRISM-Test tends to be more adversarial than real-world deployment scenarios, due to aggressive randomization of camera pose, object configuration, and more, which aids transfer.


\subsection{Extension to bimanual grasping}
\label{sec:bimanual}
While single-arm grasping has dominated robotic manipulation research, it faces inherent limitations for tasks requiring coordinated forces. Task-semantic bimanual grasping addresses these constraints by enabling opposing force application—essential for activities like unscrewing a water bottle cap, folding clothes, or lifting large objects. Our preliminary experiments in this area show that \ours{} preserves single-arm generalization capabilities while enabling task-semantic reasoning through prompt modifications. A qualitative sample may be seem in Fig.~\ref{fig:qualitative}A - while limited, this demonstrates that \ours{} supports new directions in robotic manipulation. We defer additional details to Appendix~\ref{appendix:bimanual_details}.

\section{Conclusion}
\label{sec:conclusion}

Our work demonstrates significant advances in robotic grasping capabilities, with performance improvement on nontrival task semantics and zero-shot generalization to complex real scenes. By developing a system that understands not just what an object is, but how it should be manipulated for specific intended uses, we aim to advance robotic manipulation beyond simple pick-and-place operations. Our approach demonstrates superior performance on realistic scenarios, successfully navigating busy environments and complex language instructions like ``mince some garlic" rather than the constrained ``grasp [noun] to [verb]'' paradigm of previous work. The zero-shot generalization to real scenes validates our synthetic training methodology, while our zero-shot success with bimanual grasping showcases the flexibility of our task-semantic understanding. We release \ours{}, \ourtrainingdata, \ourtestdata, and \ourrealdata, as well as our code for generating synthetic data to accelerate expansion in this direction. These contributions establish a new foundation for robotic manipulation by demonstrating a generalizable approach that functions effectively in complex and realistic environments while showing promising potential for zero-shot extension to bimanual tasks—a crucial step toward truly effective manipulation in unstructured real-world settings.

\clearpage

\section{Limitations}
\label{sec:limitations}

Despite the advancements presented in this work, several limitations remain. Our approach still maintains a dependency on a grasp proposer, though it has successfully eliminated reliances on other auxiliary models such as SAM2, GroundingDINO, and GPT-4o. Additionally, \ours{} requires integration with motion planning algorithms or alternative policies to execute agent motion effectively. Furthermore, the point-based representation may prove inadequate for scenarios demanding fine-resolution rotational adjustments in grasping operations. Future research should focus on developing models capable of directly generating semantically-appropriate stable grasps without intermediate representations.
\acknowledgments{
    We thank our coworkers at PRIOR, particularly Rohun Tripathi, for helpful discussion and feedback.
}


\clearpage
\bibliography{main}  

\begin{thebibliography}{54}
\providecommand{\natexlab}[1]{#1}
\providecommand{\url}[1]{\texttt{#1}}
\expandafter\ifx\csname urlstyle\endcsname\relax
  \providecommand{\doi}[1]{doi: #1}\else
  \providecommand{\doi}{doi: \begingroup \urlstyle{rm}\Url}\fi

\bibitem[Detry et~al.(2017)Detry, Papon, and Matthies]{detry2017task}
R.~Detry, J.~Papon, and L.~Matthies.
\newblock Task-oriented grasping with semantic and geometric scene
  understanding.
\newblock In \emph{IEEE/RSJ Intl. Conf. on Intelligent Robots and Systems
  (IROS)}, 2017.

\bibitem[Murali et~al.(2020)Murali, Liu, Marino, Chernova, and
  Gupta]{murali2020taskgrasp}
A.~Murali, W.~Liu, K.~Marino, S.~Chernova, and A.~Gupta.
\newblock Same object, different grasps: Data and semantic knowledge for
  task-oriented grasping.
\newblock In \emph{Conference on Robot Learning (CoRL)}, 2020.

\bibitem[Fang et~al.(2018)Fang, Zhu, Garg, Kurenkov, Mehta, Fei-Fei, and
  Savarese]{fang2018tog}
K.~Fang, Y.~Zhu, A.~Garg, A.~Kurenkov, V.~Mehta, L.~Fei-Fei, and S.~Savarese.
\newblock Learning task-oriented grasping for tool manipulation from simulated
  self-supervision.
\newblock In \emph{Robotics: Science and Systems (RSS)}, 2018.

\bibitem[Tang et~al.(2023)Tang, Huang, Meng, Liu, and Zhang]{tang2023graspclip}
C.~Tang, D.~Huang, L.~Meng, W.~Liu, and H.~Zhang.
\newblock Task-oriented grasp prediction with visual-language inputs.
\newblock In \emph{IEEE/RSJ Intl. Conf. on Intelligent Robots and Systems
  (IROS)}, 2023.

\bibitem[Deitke et~al.(2024)Deitke, Clark, Lee, Tripathi, Yang, Park, Salehi,
  Muennighoff, Lo, Soldaini, Lu, Anderson, Bransom, Ehsani, Ngo, Chen, Patel,
  Yatskar, Callison-Burch, Head, Hendrix, Bastani, VanderBilt, Lambert, Chou,
  Chheda, Sparks, Skjonsberg, Schmitz, Sarnat, Bischoff, Walsh, Newell,
  Wolters, Gupta, Zeng, Borchardt, Groeneveld, Nam, Lebrecht, Wittlif,
  Schoenick, Michel, Krishna, Weihs, Smith, Hajishirzi, Girshick, Farhadi, and
  Kembhavi]{deitke2024molmopixmoopenweights}
M.~Deitke, C.~Clark, S.~Lee, R.~Tripathi, Y.~Yang, J.~S. Park, M.~Salehi,
  N.~Muennighoff, K.~Lo, L.~Soldaini, J.~Lu, T.~Anderson, E.~Bransom,
  K.~Ehsani, H.~Ngo, Y.~Chen, A.~Patel, M.~Yatskar, C.~Callison-Burch, A.~Head,
  R.~Hendrix, F.~Bastani, E.~VanderBilt, N.~Lambert, Y.~Chou, A.~Chheda,
  J.~Sparks, S.~Skjonsberg, M.~Schmitz, A.~Sarnat, B.~Bischoff, P.~Walsh,
  C.~Newell, P.~Wolters, T.~Gupta, K.-H. Zeng, J.~Borchardt, D.~Groeneveld,
  C.~Nam, S.~Lebrecht, C.~Wittlif, C.~Schoenick, O.~Michel, R.~Krishna,
  L.~Weihs, N.~A. Smith, H.~Hajishirzi, R.~Girshick, A.~Farhadi, and
  A.~Kembhavi.
\newblock Molmo and pixmo: Open weights and open data for state-of-the-art
  vision-language models, 2024.
\newblock URL \url{https://arxiv.org/abs/2409.17146}.

\bibitem[Tang et~al.(2023)Tang, Huang, Ge, Liu, and Zhang]{Tang2023GraspGPTLSA}
C.~Tang, D.~Huang, W.~Ge, W.~Liu, and H.~Zhang.
\newblock Graspgpt: Leveraging semantic knowledge from a large language model
  for task-oriented grasping.
\newblock \emph{IEEE Robotics and Automation Letters}, 8:\penalty0 7551--7558,
  2023.
\newblock URL \url{https://api.semanticscholar.org/CorpusId:260154903}.

\bibitem[Tang et~al.(2024)Tang, Dong, Xu, Zhang, and
  Huang]{Tang2024FoundationGraspGTA}
C.~Tang, W.~Dong, R.~Xu, H.~Zhang, and D.~Huang.
\newblock Foundationgrasp: Generalizable task-oriented grasping with foundation
  models.
\newblock \emph{IEEE Transactions on Automation Science and Engineering},
  22:\penalty0 12418--12435, 2024.
\newblock URL \url{https://api.semanticscholar.org/CorpusId:269157340}.

\bibitem[Kerr et~al.(2023)Kerr, Kim, Goldberg, Chen, Sharma, Rashid, and
  Kanazawa]{Kerr2023LanguageERA}
J.~Kerr, C.~M. Kim, K.~Goldberg, L.~Y. Chen, S.~Sharma, A.~Rashid, and
  A.~Kanazawa.
\newblock Language embedded radiance fields for zero-shot task-oriented
  grasping.
\newblock In \emph{Conference on Robot Learning}, 2023.
\newblock URL \url{https://api.semanticscholar.org/CorpusId:261882332}.

\bibitem[Dong et~al.(2024)Dong, Huang, Liu, Tang, and Zhang]{dong2024rtagrasp}
W.~Dong, D.~Huang, J.~Liu, C.~Tang, and H.~Zhang.
\newblock Rtagrasp: Learning task-oriented grasping from human videos via
  retrieval, transfer, and alignment.
\newblock \emph{arXiv preprint arXiv:2409.16033}, 2024.

\bibitem[Tziafas et~al.(2023)Tziafas, Yucheng, Goel, Kasaei, Li, and
  Kasaei]{tziafas2023language}
G.~Tziafas, X.~Yucheng, A.~Goel, M.~Kasaei, Z.~Li, and H.~Kasaei.
\newblock Language-guided robot grasping: Clip-based referring grasp synthesis
  in clutter.
\newblock In \emph{7th Annual Conference on Robot Learning}, 2023.

\bibitem[Tang et~al.(2023)Tang, Huang, Meng, Liu, and
  Zhang]{Tang2023TaskOrientedGPA}
C.~Tang, D.~Huang, L.~Meng, W.~Liu, and H.~Zhang.
\newblock Task-oriented grasp prediction with visual-language inputs.
\newblock \emph{2023 IEEE/RSJ International Conference on Intelligent Robots
  and Systems (IROS)}, pages 4881--4888, 2023.
\newblock URL
  \url{http://ieeexplore.ieee.org/stamp/stamp.jsp?tp=\&arnumber=10342268}.

\bibitem[Xie et~al.(2024)Xie, Chen, Hu, Dai, Yang, and Wang]{xie2024target}
P.~Xie, S.~Chen, D.~Hu, Y.~Dai, K.~Yang, and G.~Wang.
\newblock Target-oriented object grasping via multimodal human guidance.
\newblock \emph{arXiv preprint arXiv:2408.11138}, 2024.

\bibitem[Liu et~al.(2023)Liu, Li, Wu, and Lee]{liu2023llava}
H.~Liu, C.~Li, Q.~Wu, and Y.~J. Lee.
\newblock Visual instruction tuning.
\newblock In \emph{NeurIPS}, 2023.

\bibitem[Achiam et~al.(2023)Achiam, Adler, Agarwal, Ahmad, Akkaya, Aleman,
  Almeida, Altenschmidt, Altman, Anadkat, et~al.]{achiam2023gpt}
J.~Achiam, S.~Adler, S.~Agarwal, L.~Ahmad, I.~Akkaya, F.~L. Aleman, D.~Almeida,
  J.~Altenschmidt, S.~Altman, S.~Anadkat, et~al.
\newblock Gpt-4 technical report.
\newblock \emph{arXiv preprint arXiv:2303.08774}, 2023.

\bibitem[Reid et~al.(2024)Reid, Savinov, Teplyashin, Lepikhin, Lillicrap,
  Alayrac, Soricut, Lazaridou, Firat, Schrittwieser, et~al.]{reid2024gemini}
M.~Reid, N.~Savinov, D.~Teplyashin, D.~Lepikhin, T.~Lillicrap, J.-b. Alayrac,
  R.~Soricut, A.~Lazaridou, O.~Firat, J.~Schrittwieser, et~al.
\newblock Gemini 1.5: Unlocking multimodal understanding across millions of
  tokens of context.
\newblock \emph{arXiv preprint arXiv:2403.05530}, 2024.

\bibitem[Zitkovich et~al.(2023)Zitkovich, Yu, Xu, Xu, Xiao, Xia, Wu, Wohlhart,
  Welker, Wahid, Vuong, Vanhoucke, Tran, Soricut, Singh, Singh, Sermanet,
  Sanketi, Salazar, Ryoo, Reymann, Rao, Pertsch, Mordatch, Michalewski, Lu,
  Levine, Lee, Lee, Leal, Kuang, Kalashnikov, Julian, Joshi, Irpan, Ichter,
  Hsu, Herzog, Hausman, Gopalakrishnan, Fu, Florence, Finn, Dubey, Driess,
  Ding, Choromanski, Chen, Chebotar, Carbajal, Brown, Brohan, Arenas, and
  Han]{zitkovich2023rt2}
B.~Zitkovich, T.~Yu, S.~Xu, P.~Xu, T.~Xiao, F.~Xia, J.~Wu, P.~Wohlhart,
  S.~Welker, A.~Wahid, Q.~Vuong, V.~Vanhoucke, H.~T. Tran, R.~Soricut,
  A.~Singh, J.~Singh, P.~Sermanet, P.~R. Sanketi, G.~Salazar, M.~S. Ryoo,
  K.~Reymann, K.~Rao, K.~Pertsch, I.~Mordatch, H.~Michalewski, Y.~Lu,
  S.~Levine, L.~Lee, T.~E. Lee, I.~Leal, Y.~Kuang, D.~Kalashnikov, R.~Julian,
  N.~J. Joshi, A.~Irpan, B.~Ichter, J.~Hsu, A.~Herzog, K.~Hausman,
  K.~Gopalakrishnan, C.~Fu, P.~Florence, C.~Finn, K.~A. Dubey, D.~Driess,
  T.~Ding, K.~M. Choromanski, X.~Chen, Y.~Chebotar, J.~Carbajal, N.~Brown,
  A.~Brohan, M.~G. Arenas, and K.~Han.
\newblock {RT-2:} vision-language-action models transfer web knowledge to
  robotic control.
\newblock In J.~Tan, M.~Toussaint, and K.~Darvish, editors, \emph{Conference on
  Robot Learning, CoRL 2023, 6-9 November 2023, Atlanta, GA, {USA}}, volume 229
  of \emph{Proceedings of Machine Learning Research}, pages 2165--2183. {PMLR},
  2023.
\newblock URL \url{https://proceedings.mlr.press/v229/zitkovich23a.html}.

\bibitem[Kim et~al.(2024)Kim, Pertsch, Karamcheti, Xiao, Balakrishna, Nair,
  Rafailov, Foster, Sanketi, Vuong, Kollar, Burchfiel, Tedrake, Sadigh, Levine,
  Liang, and Finn]{kim2024openvla}
M.~J. Kim, K.~Pertsch, S.~Karamcheti, T.~Xiao, A.~Balakrishna, S.~Nair,
  R.~Rafailov, E.~P. Foster, P.~R. Sanketi, Q.~Vuong, T.~Kollar, B.~Burchfiel,
  R.~Tedrake, D.~Sadigh, S.~Levine, P.~Liang, and C.~Finn.
\newblock Openvla: An open-source vision-language-action model.
\newblock In P.~Agrawal, O.~Kroemer, and W.~Burgard, editors, \emph{Conference
  on Robot Learning, 6-9 November 2024, Munich, Germany}, volume 270 of
  \emph{Proceedings of Machine Learning Research}, pages 2679--2713. {PMLR},
  2024.
\newblock URL \url{https://proceedings.mlr.press/v270/kim25c.html}.

\bibitem[Team et~al.(2025)Team, Abeyruwan, Ainslie, Alayrac, Arenas, Armstrong,
  Balakrishna, Baruch, Bauz{\'{a}}, Blokzijl, Bohez, Bousmalis, Brohan,
  Buschmann, Byravan, Cabi, Caluwaerts, Casarini, Chang, Chen, Chen, Chiang,
  Choromanski, D'Ambrosio, Dasari, Davchev, Devin, Palo, Ding, Dostmohamed,
  Driess, Du, Dwibedi, Elabd, Fantacci, Fong, Frey, Fu, Giustina,
  Gopalakrishnan, Graesser, Hasenclever, Heess, Hernaez, Herzog, Hofer,
  Humplik, Iscen, Jacob, Jain, Julian, Kalashnikov, Karagozler, Karp, Kew,
  Kirkland, Kirmani, Kuang, Lampe, Laurens, Leal, Lee, Lee, Liang, Lin,
  Maddineni, Majumdar, Michaely, Moreno, Neunert, Nori, Parada, Parisotto,
  Pastor, Pooley, Rao, Reymann, Sadigh, Saliceti, Sanketi, Sermanet, Shah,
  Sharma, Shea, Shu, Sindhwani, Singh, Soricut, Springenberg, Sterneck,
  Surdulescu, Tan, Tompson, Vanhoucke, Varley, Vesom, Vezzani, Vinyals, Wahid,
  and Welker]{gemini2025robotics}
G.~R. Team, S.~Abeyruwan, J.~Ainslie, J.~Alayrac, M.~G. Arenas, T.~Armstrong,
  A.~Balakrishna, R.~Baruch, M.~Bauz{\'{a}}, M.~Blokzijl, S.~Bohez,
  K.~Bousmalis, A.~Brohan, T.~Buschmann, A.~Byravan, S.~Cabi, K.~Caluwaerts,
  F.~Casarini, O.~Chang, J.~E. Chen, X.~Chen, H.~L. Chiang, K.~Choromanski,
  D.~D'Ambrosio, S.~Dasari, T.~Davchev, C.~Devin, N.~D. Palo, T.~Ding,
  A.~Dostmohamed, D.~Driess, Y.~Du, D.~Dwibedi, M.~Elabd, C.~Fantacci, C.~Fong,
  E.~Frey, C.~Fu, M.~Giustina, K.~Gopalakrishnan, L.~Graesser, L.~Hasenclever,
  N.~Heess, B.~Hernaez, A.~Herzog, R.~A. Hofer, J.~Humplik, A.~Iscen, M.~G.
  Jacob, D.~Jain, R.~Julian, D.~Kalashnikov, M.~E. Karagozler, S.~Karp, J.~C.
  Kew, J.~Kirkland, S.~Kirmani, Y.~Kuang, T.~Lampe, A.~Laurens, I.~Leal, A.~X.
  Lee, T.~E. Lee, J.~Liang, Y.~Lin, S.~Maddineni, A.~Majumdar, A.~H. Michaely,
  R.~Moreno, M.~Neunert, F.~Nori, C.~Parada, E.~Parisotto, P.~Pastor,
  A.~Pooley, K.~Rao, K.~Reymann, D.~Sadigh, S.~Saliceti, P.~Sanketi,
  P.~Sermanet, D.~Shah, M.~Sharma, K.~Shea, C.~Shu, V.~Sindhwani, S.~Singh,
  R.~Soricut, J.~T. Springenberg, R.~Sterneck, R.~Surdulescu, J.~Tan,
  J.~Tompson, V.~Vanhoucke, J.~Varley, G.~Vesom, G.~Vezzani, O.~Vinyals,
  A.~Wahid, and S.~Welker.
\newblock Gemini robotics: Bringing {AI} into the physical world.
\newblock \emph{CoRR}, abs/2503.20020, 2025.
\newblock \doi{10.48550/ARXIV.2503.20020}.
\newblock URL \url{https://doi.org/10.48550/arXiv.2503.20020}.

\bibitem[Black et~al.(2024)Black, Brown, Driess, Esmail, Equi, Finn, Fusai,
  Groom, Hausman, Ichter, Jakubczak, Jones, Ke, Levine, Li{-}Bell, Mothukuri,
  Nair, Pertsch, Shi, Tanner, Vuong, Walling, Wang, and
  Zhilinsky]{black2024pi0}
K.~Black, N.~Brown, D.~Driess, A.~Esmail, M.~Equi, C.~Finn, N.~Fusai, L.~Groom,
  K.~Hausman, B.~Ichter, S.~Jakubczak, T.~Jones, L.~Ke, S.~Levine,
  A.~Li{-}Bell, M.~Mothukuri, S.~Nair, K.~Pertsch, L.~X. Shi, J.~Tanner,
  Q.~Vuong, A.~Walling, H.~Wang, and U.~Zhilinsky.
\newblock {\(\pi\)}\({}_{\mbox{0}}\): {A} vision-language-action flow model for
  general robot control.
\newblock \emph{CoRR}, abs/2410.24164, 2024.
\newblock \doi{10.48550/ARXIV.2410.24164}.
\newblock URL \url{https://doi.org/10.48550/arXiv.2410.24164}.

\bibitem[Bjorck et~al.(2025)Bjorck, Casta{\~{n}}eda, Cherniadev, Da, Ding,
  Linxi, Fang, Fox, Hu, Huang, Jang, Jiang, Kautz, Kundalia, Lao, Li, Lin, Lin,
  Liu, LLontop, Magne, Mandlekar, Narayan, Nasiriany, Reed, Tan, Wang, Wang,
  Wang, Wang, Xiang, Xie, Xu, Xu, Ye, Yu, Zhang, Zhang, Zhao, Zheng, and
  Zhu]{bjorck2025gr00tn1}
J.~Bjorck, F.~Casta{\~{n}}eda, N.~Cherniadev, X.~Da, R.~Ding, Linxi, Y.~Fang,
  D.~Fox, F.~Hu, S.~Huang, J.~Jang, Z.~Jiang, J.~Kautz, K.~Kundalia, L.~Lao,
  Z.~Li, Z.~Lin, K.~Lin, G.~Liu, E.~LLontop, L.~Magne, A.~Mandlekar,
  A.~Narayan, S.~Nasiriany, S.~Reed, Y.~L. Tan, G.~Wang, Z.~Wang, J.~Wang,
  Q.~Wang, J.~Xiang, Y.~Xie, Y.~Xu, Z.~Xu, S.~Ye, Z.~Yu, A.~Zhang, H.~Zhang,
  Y.~Zhao, R.~Zheng, and Y.~Zhu.
\newblock {GR00T} {N1:} an open foundation model for generalist humanoid
  robots.
\newblock \emph{CoRR}, abs/2503.14734, 2025.
\newblock \doi{10.48550/ARXIV.2503.14734}.
\newblock URL \url{https://doi.org/10.48550/arXiv.2503.14734}.

\bibitem[Liu et~al.(2024)Liu, Fang, Abbeel, and Levine]{liu2024moka}
F.~Liu, K.~Fang, P.~Abbeel, and S.~Levine.
\newblock Moka: Open-vocabulary robotic manipulation through mark-based visual
  prompting.
\newblock \emph{arXiv preprint arXiv:2403.03174}, 2024.

\bibitem[Huang et~al.(2024{\natexlab{a}})Huang, Lin, Hu, Wang, and
  Gao]{huang2024copa}
H.~Huang, F.~Lin, Y.~Hu, S.~Wang, and Y.~Gao.
\newblock Copa: General robotic manipulation through spatial constraints of
  parts with foundation models.
\newblock \emph{arXiv preprint arXiv:2403.08248}, 2024{\natexlab{a}}.

\bibitem[Huang et~al.(2024{\natexlab{b}})Huang, Wang, Li, Zhang, and
  Fei-Fei]{huang2024rekep}
W.~Huang, C.~Wang, Y.~Li, R.~Zhang, and L.~Fei-Fei.
\newblock Rekep: Spatio-temporal reasoning of relational keypoint constraints
  for robotic manipulation.
\newblock \emph{arXiv preprint arXiv:2409.01652}, 2024{\natexlab{b}}.

\bibitem[Duan et~al.(2024)Duan, Yuan, Pumacay, Wang, Ehsani, Fox, and
  Krishna]{duan2024manipulate}
J.~Duan, W.~Yuan, W.~Pumacay, Y.~R. Wang, K.~Ehsani, D.~Fox, and R.~Krishna.
\newblock Manipulate-anything: Automating real-world robots using
  vision-language models.
\newblock \emph{arXiv preprint arXiv:2406.18915}, 2024.

\bibitem[Li et~al.(2024)Li, Mata, Park, Kahatapitiya, Jang, Shang, Ranasinghe,
  Burgert, Cai, Lee, et~al.]{li2024llara}
X.~Li, C.~Mata, J.~Park, K.~Kahatapitiya, Y.~S. Jang, J.~Shang, K.~Ranasinghe,
  R.~Burgert, M.~Cai, Y.~J. Lee, et~al.
\newblock Llara: Supercharging robot learning data for vision-language policy.
\newblock \emph{arXiv preprint arXiv:2406.20095}, 2024.

\bibitem[Chow et~al.(2025)Chow, Mao, Li, Seita, Guizilini, and
  Wang]{chow2025physbench}
W.~Chow, J.~Mao, B.~Li, D.~Seita, V.~Guizilini, and Y.~Wang.
\newblock Physbench: Benchmarking and enhancing vision-language models for
  physical world understanding.
\newblock \emph{arXiv preprint arXiv:2501.16411}, 2025.

\bibitem[Zhang et~al.(2024)Zhang, Huang, Ray, and Ohn-Bar]{Zhang_2024_CVPR}
J.~Zhang, Z.~Huang, A.~Ray, and E.~Ohn-Bar.
\newblock Feedback-guided autonomous driving.
\newblock In \emph{Proceedings of the IEEE/CVF Conference on Computer Vision
  and Pattern Recognition (CVPR)}, pages 15000--15011, June 2024.

\bibitem[Ehsani et~al.(2024)Ehsani, Gupta, Hendrix, Salvador, Weihs, Zeng,
  Singh, Kim, Han, Herrasti, Krishna, Schwenk, VanderBilt, and
  Kembhavi]{ehsani2024spocimitatingshortestpaths}
K.~Ehsani, T.~Gupta, R.~Hendrix, J.~Salvador, L.~Weihs, K.-H. Zeng, K.~P.
  Singh, Y.~Kim, W.~Han, A.~Herrasti, R.~Krishna, D.~Schwenk, E.~VanderBilt,
  and A.~Kembhavi.
\newblock Spoc: Imitating shortest paths in simulation enables effective
  navigation and manipulation in the real world, 2024.
\newblock URL \url{https://arxiv.org/abs/2312.02976}.

\bibitem[Yuan et~al.(2024)Yuan, Duan, Blukis, Pumacay, Krishna, Murali,
  Mousavian, and Fox]{yuan2024robopoint}
W.~Yuan, J.~Duan, V.~Blukis, W.~Pumacay, R.~Krishna, A.~Murali, A.~Mousavian,
  and D.~Fox.
\newblock Robopoint: A vision-language model for spatial affordance prediction
  for robotics.
\newblock \emph{arXiv preprint arXiv:2406.10721}, 2024.

\bibitem[Ray et~al.(2024)Ray, Duan, Tan, Bashkirova, Hendrix, Ehsani, Kembhavi,
  Plummer, Krishna, Zeng, et~al.]{ray2024sat}
A.~Ray, J.~Duan, R.~Tan, D.~Bashkirova, R.~Hendrix, K.~Ehsani, A.~Kembhavi,
  B.~A. Plummer, R.~Krishna, K.-H. Zeng, et~al.
\newblock Sat: Spatial aptitude training for multimodal language models.
\newblock \emph{arXiv preprint arXiv:2412.07755}, 2024.

\bibitem[Song et~al.(2025)Song, Blukis, Tremblay, Tyree, Su, and
  Birchfield]{song2025robospatialteachingspatialunderstanding}
C.~H. Song, V.~Blukis, J.~Tremblay, S.~Tyree, Y.~Su, and S.~Birchfield.
\newblock Robospatial: Teaching spatial understanding to 2d and 3d
  vision-language models for robotics, 2025.
\newblock URL \url{https://arxiv.org/abs/2411.16537}.

\bibitem[Brohan et~al.(2023)Brohan, Brown, Carbajal, Chebotar, Chen,
  Choromanski, Ding, Driess, Dubey, Finn, et~al.]{brohan2023rt}
A.~Brohan, N.~Brown, J.~Carbajal, Y.~Chebotar, X.~Chen, K.~Choromanski,
  T.~Ding, D.~Driess, A.~Dubey, C.~Finn, et~al.
\newblock Rt-2: Vision-language-action models transfer web knowledge to robotic
  control.
\newblock \emph{arXiv preprint arXiv:2307.15818}, 2023.

\bibitem[Zeng et~al.(2024)Zeng, Zhang, Ehsani, Hendrix, Salvador, Herrasti,
  Girshick, Kembhavi, and Weihs]{zeng2024poliformerscalingonpolicyrl}
K.-H. Zeng, Z.~Zhang, K.~Ehsani, R.~Hendrix, J.~Salvador, A.~Herrasti,
  R.~Girshick, A.~Kembhavi, and L.~Weihs.
\newblock Poliformer: Scaling on-policy rl with transformers results in
  masterful navigators, 2024.
\newblock URL \url{https://arxiv.org/abs/2406.20083}.

\bibitem[Silwal et~al.(2024)Silwal, Yadav, Wu, Vakil, Majumdar, Arnaud, Chen,
  Berges, Batra, Rajeswaran, et~al.]{silwal2024we}
S.~Silwal, K.~Yadav, T.~Wu, J.~Vakil, A.~Majumdar, S.~Arnaud, C.~Chen, V.-P.
  Berges, D.~Batra, A.~Rajeswaran, et~al.
\newblock What do we learn from a large-scale study of pre-trained visual
  representations in sim and real environments?
\newblock In \emph{2024 IEEE International Conference on Robotics and
  Automation (ICRA)}, pages 17515--17521. IEEE, 2024.

\bibitem[Sundermeyer et~al.(2021)Sundermeyer, Mousavian, Triebel, and
  Fox]{sundermeyer2021contact}
M.~Sundermeyer, A.~Mousavian, R.~Triebel, and D.~Fox.
\newblock Contact-graspnet: Efficient 6-dof grasp generation in cluttered
  scenes.
\newblock In \emph{2021 IEEE International Conference on Robotics and
  Automation (ICRA)}, pages 13438--13444. IEEE, 2021.

\bibitem[Murali et~al.(2020)Murali, Mousavian, Eppner, Paxton, and
  Fox]{murali20206}
A.~Murali, A.~Mousavian, C.~Eppner, C.~Paxton, and D.~Fox.
\newblock 6-dof grasping for target-driven object manipulation in clutter.
\newblock In \emph{2020 IEEE International Conference on Robotics and
  Automation (ICRA)}, pages 6232--6238. IEEE, 2020.

\bibitem[Jiang et~al.(2021)Jiang, Zhu, Svetlik, Fang, and
  Zhu]{jiang2021synergies}
Z.~Jiang, Y.~Zhu, M.~Svetlik, K.~Fang, and Y.~Zhu.
\newblock Synergies between affordance and geometry: 6-dof grasp detection via
  implicit representations.
\newblock \emph{Robotics: science and systems}, 2021.

\bibitem[Zeng et~al.(2020)Zeng, Florence, Tompson, Welker, Chien, Attarian,
  Armstrong, Krasin, Duong, Sindhwani, et~al.]{zeng2020transporter}
A.~Zeng, P.~Florence, J.~Tompson, S.~Welker, J.~Chien, M.~Attarian,
  T.~Armstrong, I.~Krasin, D.~Duong, V.~Sindhwani, et~al.
\newblock Transporter networks: Rearranging the visual world for robotic
  manipulation.
\newblock \emph{Conference on Robot Learning}, 2020.

\bibitem[Liu et~al.(2022)Liu, Paxton, Hermans, and Fox]{liu2022structformer}
W.~Liu, C.~Paxton, T.~Hermans, and D.~Fox.
\newblock Structformer: Learning spatial structure for language-guided semantic
  rearrangement of novel objects.
\newblock In \emph{2022 International Conference on Robotics and Automation
  (ICRA)}, pages 6322--6329. IEEE, 2022.

\bibitem[Yuan et~al.(2023)Yuan, Murali, Mousavian, and Fox]{yuan2023m2t2}
W.~Yuan, A.~Murali, A.~Mousavian, and D.~Fox.
\newblock M2t2: Multi-task masked transformer for object-centric pick and
  place.
\newblock \emph{arXiv preprint arXiv:2311.00926}, 2023.

\bibitem[Do et~al.(2018)Do, Nguyen, and Reid]{do2018affordancenet}
T.-T. Do, A.~Nguyen, and I.~Reid.
\newblock Affordancenet: An end-to-end deep learning approach for object
  affordance detection.
\newblock In \emph{International Conference on Robotics and Automation (ICRA)},
  2018.

\bibitem[Florence et~al.(2018)Florence, Manuelli, and
  Tedrake]{florencemanuelli2018dense}
P.~Florence, L.~Manuelli, and R.~Tedrake.
\newblock Dense object nets: Learning dense visual object descriptors by and
  for robotic manipulation.
\newblock \emph{Conference on Robot Learning}, 2018.

\bibitem[Manuelli et~al.(2019)Manuelli, Gao, Florence, and
  Tedrake]{manuelli2019kpam}
L.~Manuelli, W.~Gao, P.~Florence, and R.~Tedrake.
\newblock kpam: Keypoint affordances for category-level robotic manipulation.
\newblock In \emph{The International Symposium of Robotics Research}, pages
  132--157. Springer, 2019.

\bibitem[Qin et~al.(2020)Qin, Fang, Zhu, Fei-Fei, and Savarese]{qin2020keto}
Z.~Qin, K.~Fang, Y.~Zhu, L.~Fei-Fei, and S.~Savarese.
\newblock Keto: Learning keypoint representations for tool manipulation.
\newblock In \emph{2020 IEEE International Conference on Robotics and
  Automation (ICRA)}, pages 7278--7285. IEEE, 2020.

\bibitem[Mo et~al.(2021)Mo, Guibas, Mukadam, Gupta, and
  Tulsiani]{mo2021where2act}
K.~Mo, L.~J. Guibas, M.~Mukadam, A.~Gupta, and S.~Tulsiani.
\newblock Where2act: From pixels to actions for articulated 3d objects.
\newblock In \emph{Proceedings of the IEEE/CVF International Conference on
  Computer Vision}, pages 6813--6823, 2021.

\bibitem[Shridhar et~al.(2022)Shridhar, Manuelli, and Fox]{shridhar2022cliport}
M.~Shridhar, L.~Manuelli, and D.~Fox.
\newblock Cliport: What and where pathways for robotic manipulation.
\newblock In \emph{Conference on robot learning}, pages 894--906. PMLR, 2022.

\bibitem[Chang et~al.(2015)Chang, Funkhouser, Guibas, Hanrahan, Huang, Li,
  Savarese, Savva, Song, Su, Xiao, Yi, and Yu]{shapenet2015}
A.~X. Chang, T.~Funkhouser, L.~Guibas, P.~Hanrahan, Q.~Huang, Z.~Li,
  S.~Savarese, M.~Savva, S.~Song, H.~Su, J.~Xiao, L.~Yi, and F.~Yu.
\newblock {ShapeNet: An Information-Rich 3D Model Repository}.
\newblock Technical Report arXiv:1512.03012 [cs.GR], Stanford University ---
  Princeton University --- Toyota Technological Institute at Chicago, 2015.

\bibitem[Eppner et~al.(2021)Eppner, Mousavian, and Fox]{acronym2020}
C.~Eppner, A.~Mousavian, and D.~Fox.
\newblock Acronym: A large-scale grasp dataset based on simulation.
\newblock In \emph{2021 IEEE International Conference on Robotics and
  Automation (ICRA)}, pages 6222--6227, 2021.
\newblock \doi{10.1109/ICRA48506.2021.9560844}.

\bibitem[Savva et~al.(2015)Savva, Chang, and Hanrahan]{savva2015semgeo}
M.~Savva, A.~X. Chang, and P.~Hanrahan.
\newblock {Semantically-Enriched 3D Models for Common-sense Knowledge}.
\newblock \emph{CVPR 2015 Workshop on Functionality, Physics, Intentionality
  and Causality}, 2015.

\bibitem[Eppner et~al.(2024)Eppner, Murali, Garrett, O'Flaherty, Hermans, Yang,
  and Fox]{Eppner2024}
C.~Eppner, A.~Murali, C.~Garrett, R.~O'Flaherty, T.~Hermans, W.~Yang, and
  D.~Fox.
\newblock scene\_synthesizer: A python library for procedural scene generation
  in robot manipulation.
\newblock \emph{Journal of Open Source Software}, 2024.

\bibitem[OpenAI et~al.(2024)OpenAI, :, Hurst, Lerer, Goucher, Perelman, Ramesh,
  Clark, Ostrow, Welihinda, Hayes, Radford, Madry, Baker-Whitcomb, Beutel,
  Borzunov, Carney, Chow, Kirillov, Nichol, Paino, Renzin, Passos, Kirillov,
  Christakis, Conneau, Kamali, Jabri, Moyer, Tam, Crookes, Tootoochian,
  Tootoonchian, Kumar, Vallone, Karpathy, Braunstein, Cann, Codispoti, Galu,
  Kondrich, Tulloch, Mishchenko, Baek, Jiang, Pelisse, Woodford, Gosalia, Dhar,
  Pantuliano, Nayak, Oliver, Zoph, Ghorbani, Leimberger, Rossen, Sokolowsky,
  Wang, Zweig, Hoover, Samic, McGrew, Spero, Giertler, Cheng, Lightcap, Walkin,
  Quinn, Guarraci, Hsu, Kellogg, Eastman, Lugaresi, Wainwright, Bassin, Hudson,
  Chu, Nelson, Li, Shern, Conger, Barette, Voss, Ding, Lu, Zhang, Beaumont,
  Hallacy, Koch, Gibson, Kim, Choi, McLeavey, Hesse, Fischer, Winter,
  Czarnecki, Jarvis, Wei, Koumouzelis, Sherburn, Kappler, Levin, Levy, Carr,
  Farhi, Mely, Robinson, Sasaki, Jin, Valladares, Tsipras, Li, Nguyen, Findlay,
  Oiwoh, Wong, Asdar, Proehl, Yang, Antonow, Kramer, Peterson, Sigler, Wallace,
  Brevdo, Mays, Khorasani, Such, Raso, Zhang, von Lohmann, Sulit, Goh, Oden,
  Salmon, Starace, Brockman, Salman, Bao, Hu, Wong, Wang, Schmidt, Whitney,
  Jun, Kirchner, de~Oliveira~Pinto, Ren, Chang, Chung, Kivlichan, O'Connell,
  O'Connell, Osband, Silber, Sohl, Okuyucu, Lan, Kostrikov, Sutskever,
  Kanitscheider, Gulrajani, Coxon, Menick, Pachocki, Aung, Betker, Crooks,
  Lennon, Kiros, Leike, Park, Kwon, Phang, Teplitz, Wei, Wolfe, Chen, Harris,
  Varavva, Lee, Shieh, Lin, Yu, Weng, Tang, Yu, Jang, Candela, Beutler,
  Landers, Parish, Heidecke, Schulman, Lachman, McKay, Uesato, Ward, Kim,
  Huizinga, Sitkin, Kraaijeveld, Gross, Kaplan, Snyder, Achiam, Jiao, Lee,
  Zhuang, Harriman, Fricke, Hayashi, Singhal, Shi, Karthik, Wood, Rimbach, Hsu,
  Nguyen, Gu-Lemberg, Button, Liu, Howe, Muthukumar, Luther, Ahmad, Kai, Itow,
  Workman, Pathak, Chen, Jing, Guy, Fedus, Zhou, Mamitsuka, Weng, McCallum,
  Held, Ouyang, Feuvrier, Zhang, Kondraciuk, Kaiser, Hewitt, Metz, Doshi,
  Aflak, Simens, Boyd, Thompson, Dukhan, Chen, Gray, Hudnall, Zhang, Aljubeh,
  Litwin, Zeng, Johnson, Shetty, Gupta, Shah, Yatbaz, Yang, Zhong, Glaese,
  Chen, Janner, Lampe, Petrov, Wu, Wang, Fradin, Pokrass, Castro, de~Castro,
  Pavlov, Brundage, Wang, Khan, Murati, Bavarian, Lin, Yesildal, Soto,
  Gimelshein, Cone, Staudacher, Summers, LaFontaine, Chowdhury, Ryder, Stathas,
  Turley, Tezak, Felix, Kudige, Keskar, Deutsch, Bundick, Puckett, Nachum,
  Okelola, Boiko, Murk, Jaffe, Watkins, Godement, Campbell-Moore, Chao,
  McMillan, Belov, Su, Bak, Bakkum, Deng, Dolan, Hoeschele, Welinder, Tillet,
  Pronin, Tillet, Dhariwal, Yuan, Dias, Lim, Arora, Troll, Lin, Lopes, Puri,
  Miyara, Leike, Gaubert, Zamani, Wang, Donnelly, Honsby, Smith, Sahai,
  Ramchandani, Huet, Carmichael, Zellers, Chen, Chen, Nigmatullin, Cheu, Jain,
  Altman, Schoenholz, Toizer, Miserendino, Agarwal, Culver, Ethersmith, Gray,
  Grove, Metzger, Hermani, Jain, Zhao, Wu, Jomoto, Wu, Shuaiqi, Xia, Phene,
  Papay, Narayanan, Coffey, Lee, Hall, Balaji, Broda, Stramer, Xu, Gogineni,
  Christianson, Sanders, Patwardhan, Cunninghman, Degry, Dimson, Raoux,
  Shadwell, Zheng, Underwood, Markov, Sherbakov, Rubin, Stasi, Kaftan, Heywood,
  Peterson, Walters, Eloundou, Qi, Moeller, Monaco, Kuo, Fomenko, Chang, Zheng,
  Zhou, Manassra, Sheu, Zaremba, Patil, Qian, Kim, Cheng, Zhang, He, Zhang,
  Jin, Dai, and Malkov]{openai2024gpt4ocard}
OpenAI, :, A.~Hurst, A.~Lerer, A.~P. Goucher, A.~Perelman, A.~Ramesh, A.~Clark,
  A.~Ostrow, A.~Welihinda, A.~Hayes, A.~Radford, A.~Madry, A.~Baker-Whitcomb,
  A.~Beutel, A.~Borzunov, A.~Carney, A.~Chow, A.~Kirillov, A.~Nichol, A.~Paino,
  A.~Renzin, A.~T. Passos, A.~Kirillov, A.~Christakis, A.~Conneau, A.~Kamali,
  A.~Jabri, A.~Moyer, A.~Tam, A.~Crookes, A.~Tootoochian, A.~Tootoonchian,
  A.~Kumar, A.~Vallone, A.~Karpathy, A.~Braunstein, A.~Cann, A.~Codispoti,
  A.~Galu, A.~Kondrich, A.~Tulloch, A.~Mishchenko, A.~Baek, A.~Jiang,
  A.~Pelisse, A.~Woodford, A.~Gosalia, A.~Dhar, A.~Pantuliano, A.~Nayak,
  A.~Oliver, B.~Zoph, B.~Ghorbani, B.~Leimberger, B.~Rossen, B.~Sokolowsky,
  B.~Wang, B.~Zweig, B.~Hoover, B.~Samic, B.~McGrew, B.~Spero, B.~Giertler,
  B.~Cheng, B.~Lightcap, B.~Walkin, B.~Quinn, B.~Guarraci, B.~Hsu, B.~Kellogg,
  B.~Eastman, C.~Lugaresi, C.~Wainwright, C.~Bassin, C.~Hudson, C.~Chu,
  C.~Nelson, C.~Li, C.~J. Shern, C.~Conger, C.~Barette, C.~Voss, C.~Ding,
  C.~Lu, C.~Zhang, C.~Beaumont, C.~Hallacy, C.~Koch, C.~Gibson, C.~Kim,
  C.~Choi, C.~McLeavey, C.~Hesse, C.~Fischer, C.~Winter, C.~Czarnecki,
  C.~Jarvis, C.~Wei, C.~Koumouzelis, D.~Sherburn, D.~Kappler, D.~Levin,
  D.~Levy, D.~Carr, D.~Farhi, D.~Mely, D.~Robinson, D.~Sasaki, D.~Jin,
  D.~Valladares, D.~Tsipras, D.~Li, D.~P. Nguyen, D.~Findlay, E.~Oiwoh,
  E.~Wong, E.~Asdar, E.~Proehl, E.~Yang, E.~Antonow, E.~Kramer, E.~Peterson,
  E.~Sigler, E.~Wallace, E.~Brevdo, E.~Mays, F.~Khorasani, F.~P. Such, F.~Raso,
  F.~Zhang, F.~von Lohmann, F.~Sulit, G.~Goh, G.~Oden, G.~Salmon, G.~Starace,
  G.~Brockman, H.~Salman, H.~Bao, H.~Hu, H.~Wong, H.~Wang, H.~Schmidt,
  H.~Whitney, H.~Jun, H.~Kirchner, H.~P. de~Oliveira~Pinto, H.~Ren, H.~Chang,
  H.~W. Chung, I.~Kivlichan, I.~O'Connell, I.~O'Connell, I.~Osband, I.~Silber,
  I.~Sohl, I.~Okuyucu, I.~Lan, I.~Kostrikov, I.~Sutskever, I.~Kanitscheider,
  I.~Gulrajani, J.~Coxon, J.~Menick, J.~Pachocki, J.~Aung, J.~Betker,
  J.~Crooks, J.~Lennon, J.~Kiros, J.~Leike, J.~Park, J.~Kwon, J.~Phang,
  J.~Teplitz, J.~Wei, J.~Wolfe, J.~Chen, J.~Harris, J.~Varavva, J.~G. Lee,
  J.~Shieh, J.~Lin, J.~Yu, J.~Weng, J.~Tang, J.~Yu, J.~Jang, J.~Q. Candela,
  J.~Beutler, J.~Landers, J.~Parish, J.~Heidecke, J.~Schulman, J.~Lachman,
  J.~McKay, J.~Uesato, J.~Ward, J.~W. Kim, J.~Huizinga, J.~Sitkin,
  J.~Kraaijeveld, J.~Gross, J.~Kaplan, J.~Snyder, J.~Achiam, J.~Jiao, J.~Lee,
  J.~Zhuang, J.~Harriman, K.~Fricke, K.~Hayashi, K.~Singhal, K.~Shi,
  K.~Karthik, K.~Wood, K.~Rimbach, K.~Hsu, K.~Nguyen, K.~Gu-Lemberg, K.~Button,
  K.~Liu, K.~Howe, K.~Muthukumar, K.~Luther, L.~Ahmad, L.~Kai, L.~Itow,
  L.~Workman, L.~Pathak, L.~Chen, L.~Jing, L.~Guy, L.~Fedus, L.~Zhou,
  L.~Mamitsuka, L.~Weng, L.~McCallum, L.~Held, L.~Ouyang, L.~Feuvrier,
  L.~Zhang, L.~Kondraciuk, L.~Kaiser, L.~Hewitt, L.~Metz, L.~Doshi, M.~Aflak,
  M.~Simens, M.~Boyd, M.~Thompson, M.~Dukhan, M.~Chen, M.~Gray, M.~Hudnall,
  M.~Zhang, M.~Aljubeh, M.~Litwin, M.~Zeng, M.~Johnson, M.~Shetty, M.~Gupta,
  M.~Shah, M.~Yatbaz, M.~J. Yang, M.~Zhong, M.~Glaese, M.~Chen, M.~Janner,
  M.~Lampe, M.~Petrov, M.~Wu, M.~Wang, M.~Fradin, M.~Pokrass, M.~Castro,
  M.~O.~T. de~Castro, M.~Pavlov, M.~Brundage, M.~Wang, M.~Khan, M.~Murati,
  M.~Bavarian, M.~Lin, M.~Yesildal, N.~Soto, N.~Gimelshein, N.~Cone,
  N.~Staudacher, N.~Summers, N.~LaFontaine, N.~Chowdhury, N.~Ryder, N.~Stathas,
  N.~Turley, N.~Tezak, N.~Felix, N.~Kudige, N.~Keskar, N.~Deutsch, N.~Bundick,
  N.~Puckett, O.~Nachum, O.~Okelola, O.~Boiko, O.~Murk, O.~Jaffe, O.~Watkins,
  O.~Godement, O.~Campbell-Moore, P.~Chao, P.~McMillan, P.~Belov, P.~Su,
  P.~Bak, P.~Bakkum, P.~Deng, P.~Dolan, P.~Hoeschele, P.~Welinder, P.~Tillet,
  P.~Pronin, P.~Tillet, P.~Dhariwal, Q.~Yuan, R.~Dias, R.~Lim, R.~Arora,
  R.~Troll, R.~Lin, R.~G. Lopes, R.~Puri, R.~Miyara, R.~Leike, R.~Gaubert,
  R.~Zamani, R.~Wang, R.~Donnelly, R.~Honsby, R.~Smith, R.~Sahai,
  R.~Ramchandani, R.~Huet, R.~Carmichael, R.~Zellers, R.~Chen, R.~Chen,
  R.~Nigmatullin, R.~Cheu, S.~Jain, S.~Altman, S.~Schoenholz, S.~Toizer,
  S.~Miserendino, S.~Agarwal, S.~Culver, S.~Ethersmith, S.~Gray, S.~Grove,
  S.~Metzger, S.~Hermani, S.~Jain, S.~Zhao, S.~Wu, S.~Jomoto, S.~Wu, Shuaiqi,
  Xia, S.~Phene, S.~Papay, S.~Narayanan, S.~Coffey, S.~Lee, S.~Hall, S.~Balaji,
  T.~Broda, T.~Stramer, T.~Xu, T.~Gogineni, T.~Christianson, T.~Sanders,
  T.~Patwardhan, T.~Cunninghman, T.~Degry, T.~Dimson, T.~Raoux, T.~Shadwell,
  T.~Zheng, T.~Underwood, T.~Markov, T.~Sherbakov, T.~Rubin, T.~Stasi,
  T.~Kaftan, T.~Heywood, T.~Peterson, T.~Walters, T.~Eloundou, V.~Qi,
  V.~Moeller, V.~Monaco, V.~Kuo, V.~Fomenko, W.~Chang, W.~Zheng, W.~Zhou,
  W.~Manassra, W.~Sheu, W.~Zaremba, Y.~Patil, Y.~Qian, Y.~Kim, Y.~Cheng,
  Y.~Zhang, Y.~He, Y.~Zhang, Y.~Jin, Y.~Dai, and Y.~Malkov.
\newblock Gpt-4o system card, 2024.
\newblock URL \url{https://arxiv.org/abs/2410.21276}.

\bibitem[Ravi et~al.(2024)Ravi, Gabeur, Hu, Hu, Ryali, Ma, Khedr, R{\"a}dle,
  Rolland, Gustafson, Mintun, Pan, Alwala, Carion, Wu, Girshick, Doll{\'a}r,
  and Feichtenhofer]{ravi2024sam2}
N.~Ravi, V.~Gabeur, Y.-T. Hu, R.~Hu, C.~Ryali, T.~Ma, H.~Khedr, R.~R{\"a}dle,
  C.~Rolland, L.~Gustafson, E.~Mintun, J.~Pan, K.~V. Alwala, N.~Carion, C.-Y.
  Wu, R.~Girshick, P.~Doll{\'a}r, and C.~Feichtenhofer.
\newblock Sam 2: Segment anything in images and videos.
\newblock \emph{arXiv preprint arXiv:2408.00714}, 2024.
\newblock URL \url{https://arxiv.org/abs/2408.00714}.

\bibitem[Liu et~al.(2023)Liu, Zeng, Ren, Li, Zhang, Yang, Li, Yang, Su, Zhu,
  et~al.]{liu2023grounding}
S.~Liu, Z.~Zeng, T.~Ren, F.~Li, H.~Zhang, J.~Yang, C.~Li, J.~Yang, H.~Su,
  J.~Zhu, et~al.
\newblock Grounding dino: Marrying dino with grounded pre-training for open-set
  object detection.
\newblock \emph{arXiv preprint arXiv:2303.05499}, 2023.

\bibitem[Yuan et~al.(2020)Yuan, Eckart, Kim, Jampani, Fox, and
  Kautz]{yuan2020deepgmr}
W.~Yuan, B.~Eckart, K.~Kim, V.~Jampani, D.~Fox, and J.~Kautz.
\newblock Deepgmr: Learning latent gaussian mixture models for registration.
\newblock In \emph{European Conference on Computer Vision}, pages 733--750.
  Springer, 2020.

\end{thebibliography}

\clearpage
\appendix
\section{Appendix}
\label{appendix}

\subsection{\ourdata{} Scene Generation}
\label{appendix:scene_generation_details}

\ourdata{} consists of 100k views of 10k scenes, generated via procedural generation of 2,365 object instances across 91 object classes, enumerated in Table~\ref{tab:object_classes}. Pocedural scene composition allows us to generate a wide variety of scenes at scale, maximizing both data diversity and quantity. We employ aggressive randomization in this process, the bounds for which are outlined in Table~\ref{tab:scene-randomization}.

To evaluate a model's ability to perform task-oriented grasping on new scenes, we create \ourtestdata{}, composed of novel object instances and classes. Specifically, we hold out 10\% (rounded up) of object instances from each object class, and also 4 entire object classes (TeaCup, Fork, DSLRCamera, and PillBottle). This not only ensures that models do not overfit to the objects seen during training, but also tests their ability to generalize grasping to completely novel objects.

\begin{table}[b]
\centering
\renewcommand{\arraystretch}{1.1}
\setlength{\tabcolsep}{0.4cm}
\begin{tabular}{lc}
    \toprule
    Min. distance from table to wall (m) & 2 \\
    Room dimensions (m$\times$m) & $[4, 10]\times[4,10]$ \\
    Camera DFOV (deg) & $[60, 110]$ \\
    Camera distance (m) & $[0.25, 1.25]$ \\
    Camera pitch perturbation (frac. of VFOV) & $[0, 0.02]$ \\
    Camera yaw perturbation (frac. of HFOV) & $[0, 0.05]$ \\
    Camera roll perturbation (rad) & $[0, 0.39]$ \\
    Camera elevation (rad) & $[\pi/8, \pi/3]$ \\
    Image size (px) & $480 \times 640$ \\
    Number of views per scene & 10 \\
    Min. \# of annotations per view & 2 \\
    Number of objects & $[6, 12]$ \\
    Max. grasp distance (m) & 1.0 \\
    Color temperature (K) & $[2000, 10000]$ \\
    Light intensity (lux) & $[10, 25]$ \\
    Light azimuth (rad) & $[0, 2\pi]$ \\
    Light inclination (rad) & $[0, \pi/3]$ \\
    \bottomrule
\end{tabular}
\smallskip
\caption{The randomization parameters used for scene generation in \ourdata{}.}
\label{tab:scene-randomization}
\end{table}

\begin{table*}[b]
\centering
\emph{
\begin{tabular}{l}
\toprule
banana, bag, beer bottle, book, bottle, bowl, bread slice, calculator, camera, candle, canister,\\ can opener, cap, carrot, cassette, cell phone, cereal box, chocolate, coaster, coin,\\ computer mouse, controller, cookie, cup, cup cake, desk lamp, disc case, donut, drink bottle,\\ drinking utensil, DSLR camera, eraser, flashlight, food item, fork, fruit, glasses, guitar, hammer,\\ hanger, hat, headphones, keyboard, knife, laptop, magnet, marker, media discs, milk carton,\\ mouse pad, mug, Nintendo DS, notepad, pan, paper box, paper clip, pen, pencil, picture frame,\\ pill bottle, plate, power strip, purse, radio, ring, Rubik's cube, ruler, scissors, screw driver,\\ shampoo, shoes, soap bar, soap bottle, soda can, spoon, stapler, table clock, table lamp,\\ tape measure, teacup, teapot, toilet paper, USB stick, video game controller, wall clock,\\ wallet, watch, web cam, Wii, wine bottle, \emph{and} wine glass\\
\bottomrule
\end{tabular}%
}
\caption{\label{tab:object_classes}The 91 \textbf{object classes} corresponding to the 2,365 objects included in \ourdata{}.}
\end{table*}

\subsection{Grasp Sampling}
\label{appendix:grasp_sampling}

The ACRONYM dataset provides roughly 2,000 grasps on each object mesh in the dataset, about half of which are labeled as stable grasps. Since we aim to collect object-grasp descriptions, it would be prohibitively expensive to do so for every grasp in the ACRONYM dataset. To alleviate this, we sample a subset of grasps for each object mesh for labeling. In practice, we sample 4 grasps per object mesh.

Standard practice for subsampling would be to use farthest-point sampling, i.e. per-instance grasp sampling. However, we see that doing so for each object mesh independently tends to create clusters of grasps, instead of creating a uniform distribution of grasps over the object. This is largely due to the fact that objects of the same class tend to have similar morphologies, and therefore have certain parts that are closer to or further from each other. This results in farthest point sampling picking grasps on similar parts for every object instance in a class, reducing the diversity of grasps on a type of object.

\begin{figure}
    \centering
    \begin{subfigure}{0.4\linewidth}
        \includegraphics[width=\linewidth]{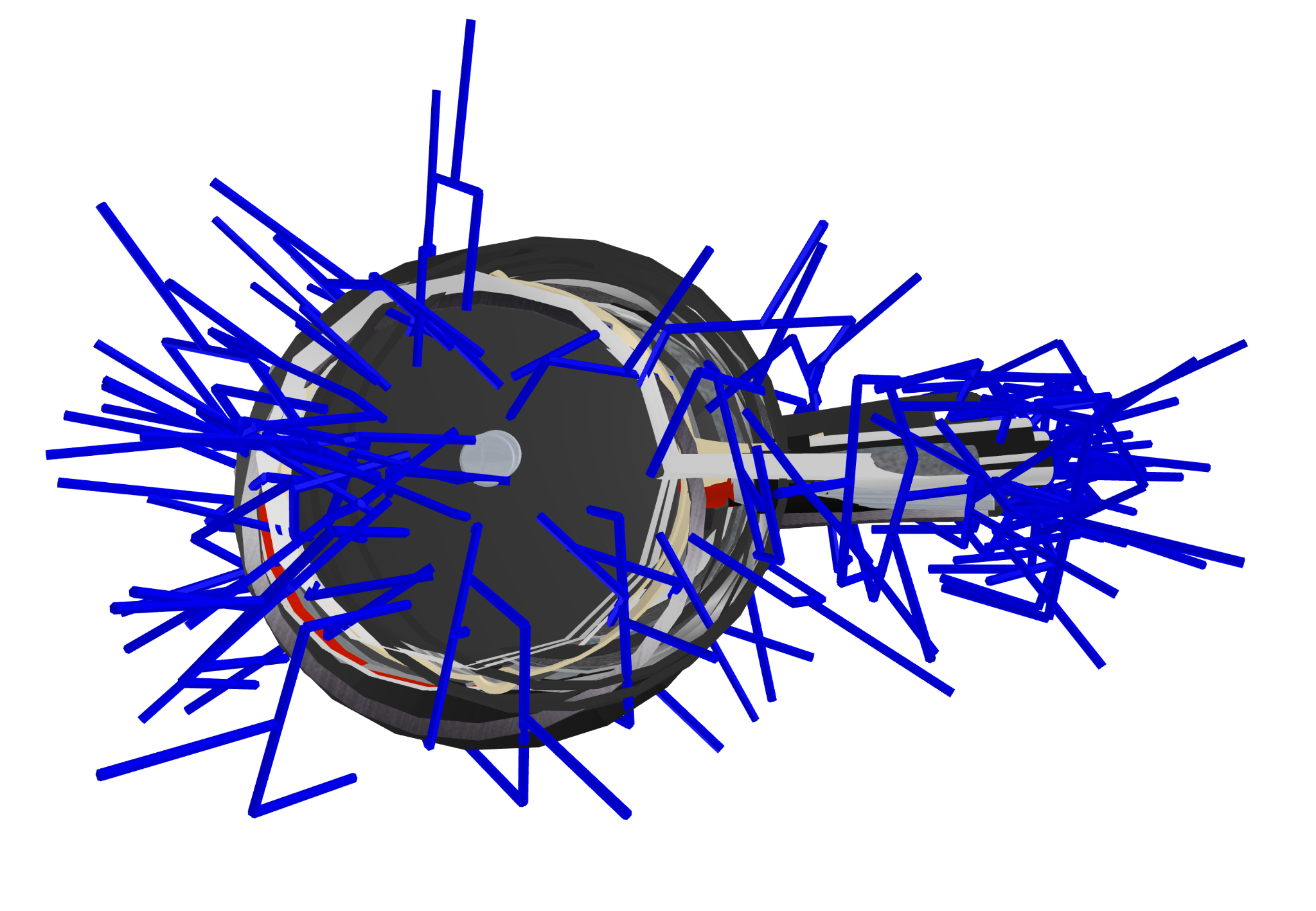}
        \caption{Per-instance grasp sampling}
    \end{subfigure}
    \begin{subfigure}{0.4\linewidth}
        \includegraphics[width=\linewidth]{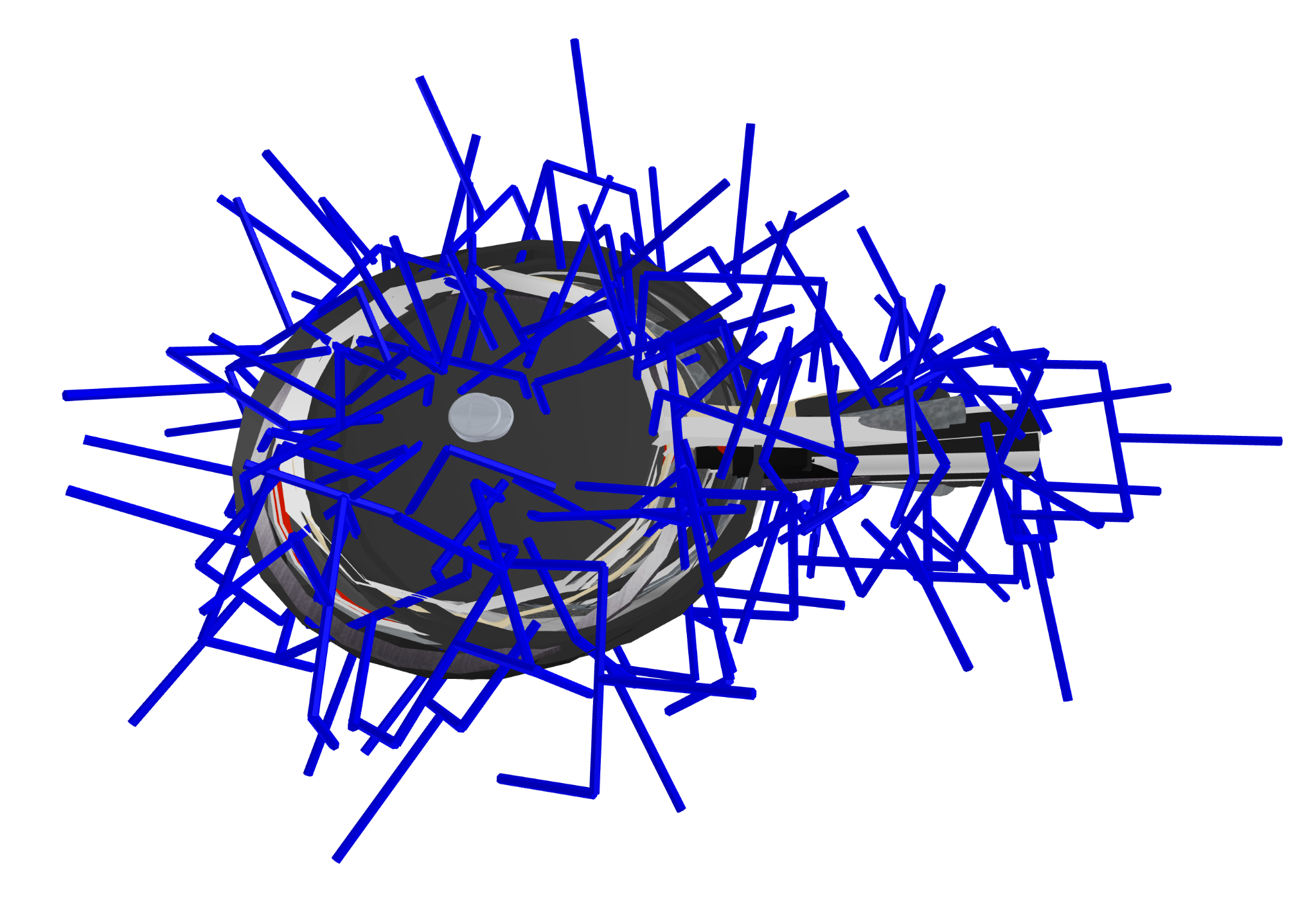}
        \caption{Cross-instance grasp sampling}
    \end{subfigure}
    \caption{Cross-instance grasp sampling (right) creates a subset of grasps with much better coverage and diversity than indendent per-instance grasp sampling (left).}
    \label{fig:grasp-sampling}
\end{figure}

To counteract this, we introduce cross-instance grasp sampling, visualized in Figure~\ref{fig:grasp-sampling}. Concretely, within a class, we sample grasps on an object instance while considering grasps on similar parts of different object instances. This can be achieved by aligning object meshes within a class to each other, which we do as follows. We first notice that since the up-vector is known for all objects meshes, we need only to align the object meshes in the xy-plane. We first project the mesh to the xy-plane, and sample 1000 points within the resulting polygon. We then align the centroids of these sets of points, and align orientations using the principal axes. We finally refine with iterative closest point (ICP) to retrieve a transform, which we apply to the object mesh.

After aligning meshes within a class, we use farthest point sampling in a round-robin fashion, sampling the furthest grasp for each object instance until we get 4 grasps per instance. For any object meshes for which this registration process fails, we use farthest point sampling independent of the other meshes.

\subsection{Grasp Description Generation}

\begin{figure}[htb]
    \centering
    \includegraphics[width=\linewidth]{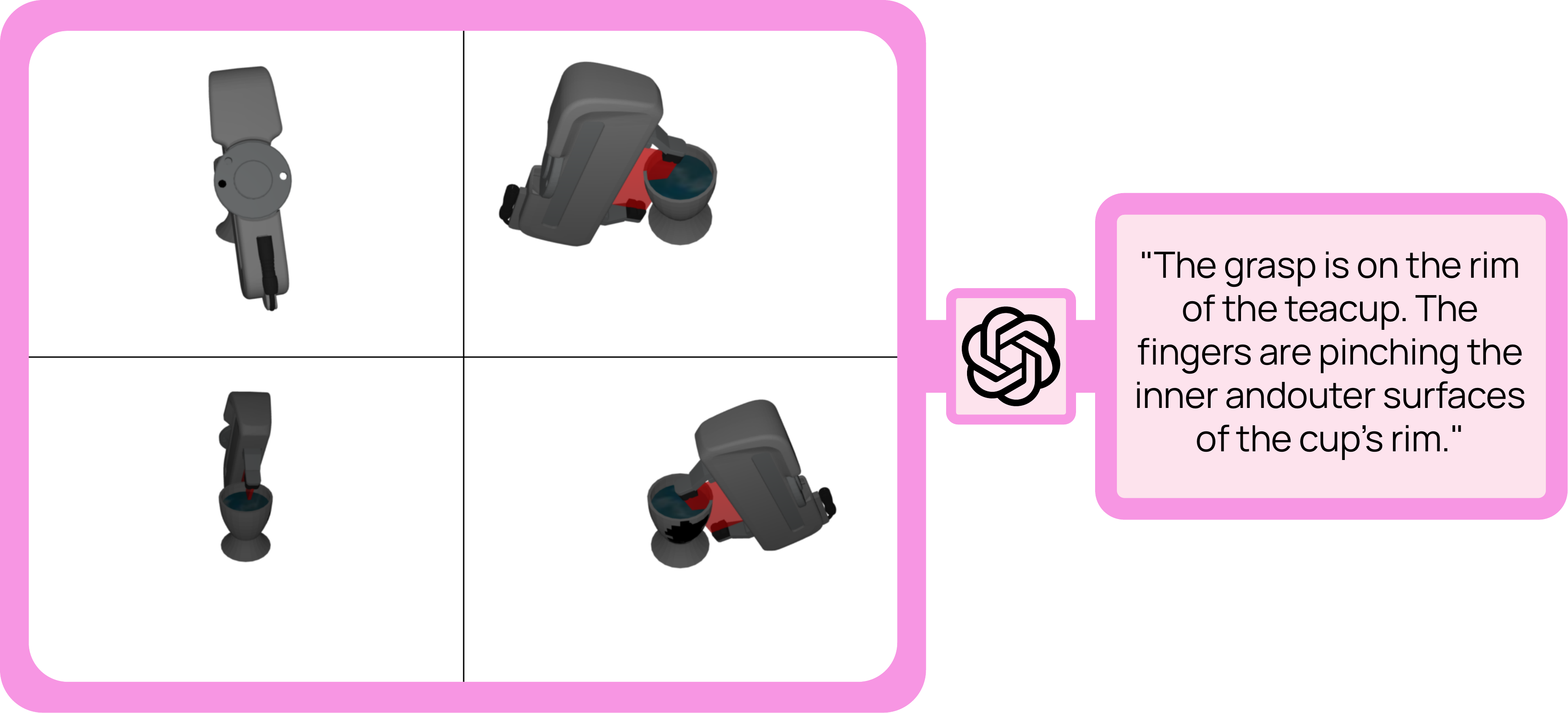}
    \caption{Given multiple views of a grasp on an object mesh, GPT-4o describes the grasp in natural language. The red rectangle rendered in the collage helps the VLM to understand the volume grasped by the gripper, and how it intersects the object mesh.}
    \label{fig:grasp-description}
\end{figure}

In Section~\ref{sec:datagen}, we outline how we use GPT-4o to generate synthetic grasp descriptions. We illustrate an example of this process in Figure~\ref{fig:grasp-description}. Since GPT-4o can hallucinate incorrect spatial relations or grounding, we ask Prolific workers to manually verify the generated descriptions, and provide corrections if necessary. Full prompts are given in Figure~\ref{fig:prompt-synthetic-grasp-description-generation}.

\subsection{Task Generation}
\label{appendix:task_generation_details}

We provide more details about the two steps involved in semantic grasping task generation (Task Generation paragraph in Section \ref{sec:datagen}).

For the \textbf{grasp description generation} step, the structure of the required response is made to facilitate a causal explanation of the choices made while discouraging hallucinations. First, the LLM generates a list of object subtypes and identifies a list of optional parts only present in some subtypes, followed by a list of common object parts for any object of the given type (and subtypes). We then let the LLM assume a relative starting pose of the object relative to the supporting surface and describe it as the object parts in contact with the surface. The LLM then generates a list of common graspable parts, excluding optional parts, and the target list of grasp descriptors. Each grasp is generated as (1) the object part to contact, (2) an example semantic task, (3) an approach direction, finger-plane and gripper orientation, which provide redundancy to discourage inconsistent or implausible grasps, and (4) a natural language description of the grasp (excluding any reference to the example task). The prompt in use is shown in Figure \ref{fig:prompt-grasp-description-generation}.

For the \textbf{semantic manipulation task generation} step we ask the LLM to generate four valid semantic tasks per grasp conditioned on the object class, its relative starting pose, and the target and alternative grasp descriptions (excluding the example semantic tasks used to guide grasp generation). Each semantic task is generated as (1) the task instruction in natural language (avoiding any reference to the grasp), complemented with the number of grippers required to complete the task beyond the initial grasp, (2) a grasp critique seeking possible unsuitability of the grasp for the task and a grasp score in the range 0 (worst validity) to 9 (best), (3) an alternative grasp critique seeking possible suitability of the alternative grasp and corresponding grasp score, and (4) an analysis of the validity of the task definition, considering lack of assumptions about the object's state and context, or physical plausibility of the grasp and task, among others, and a corresponding validity score, also in the range 0 to 9. The prompt in use is shown in Figure \ref{fig:prompt-semantic-task-generation}.

\subsection{Grasp Matching}

\begin{figure}[htb]
    \centering
    \renewcommand{\arraystretch}{5}
    \begin{tabular}{c p{0.6\linewidth}}
        \toprule
        \raisebox{-0.4\totalheight}{\includegraphics[width=2cm]{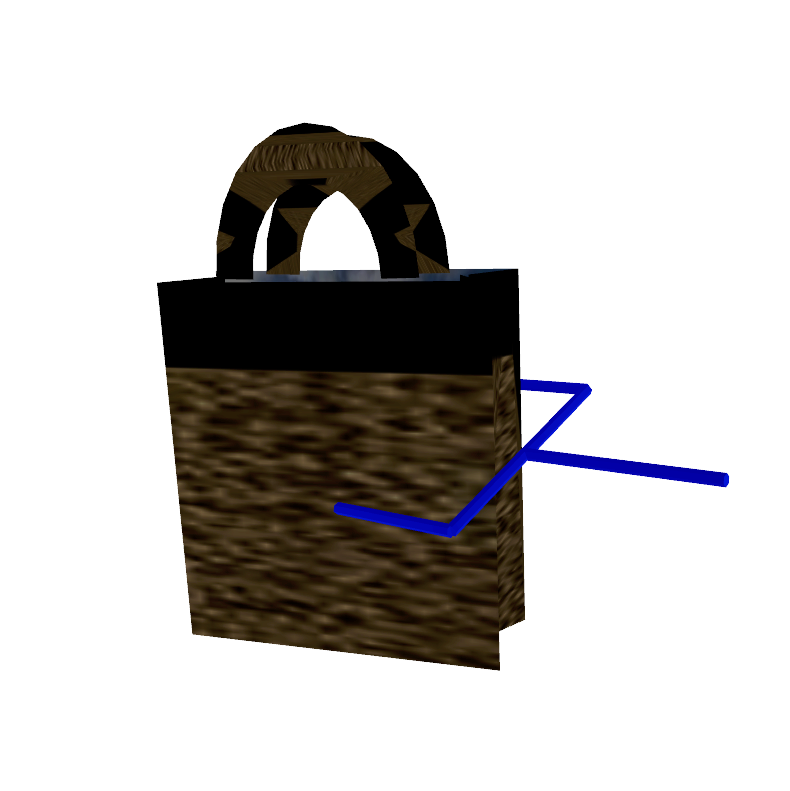}} & Gently press down on the bag’s main body to flatten it against the surface. \\
        \raisebox{-0.4\totalheight}{\includegraphics[width=2cm]{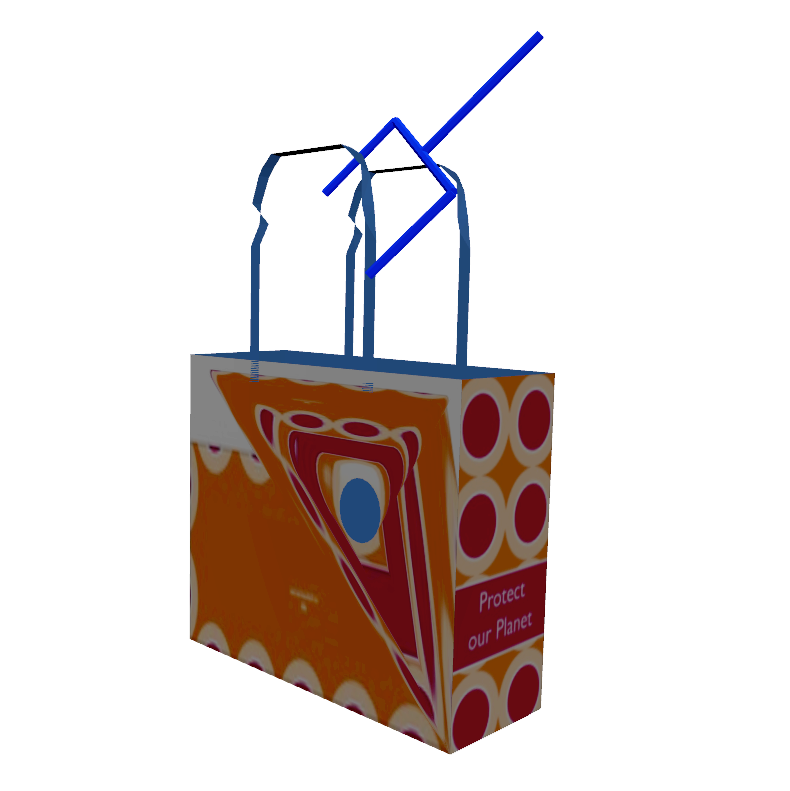}} & Lift the bag off the surface and swing it gently forward, as to set it onto a nearby bench. \\ \hline
        \raisebox{-0.4\totalheight}{\includegraphics[width=2cm]{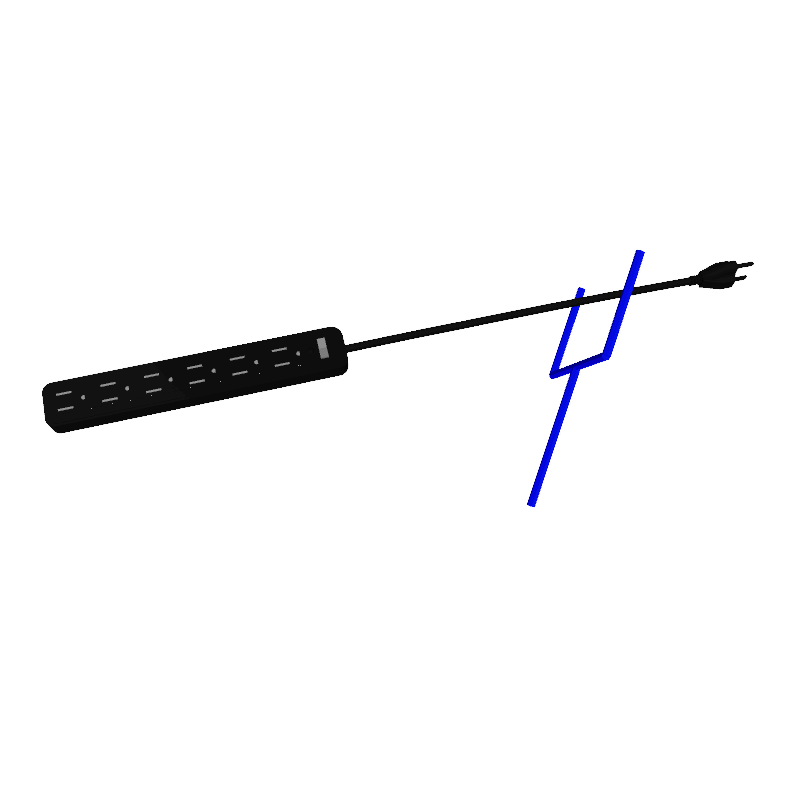}} & Feed the power strip's cord into a cord management clip affixed to the side of the table. \\
        \raisebox{-0.4\totalheight}{\includegraphics[width=2cm]{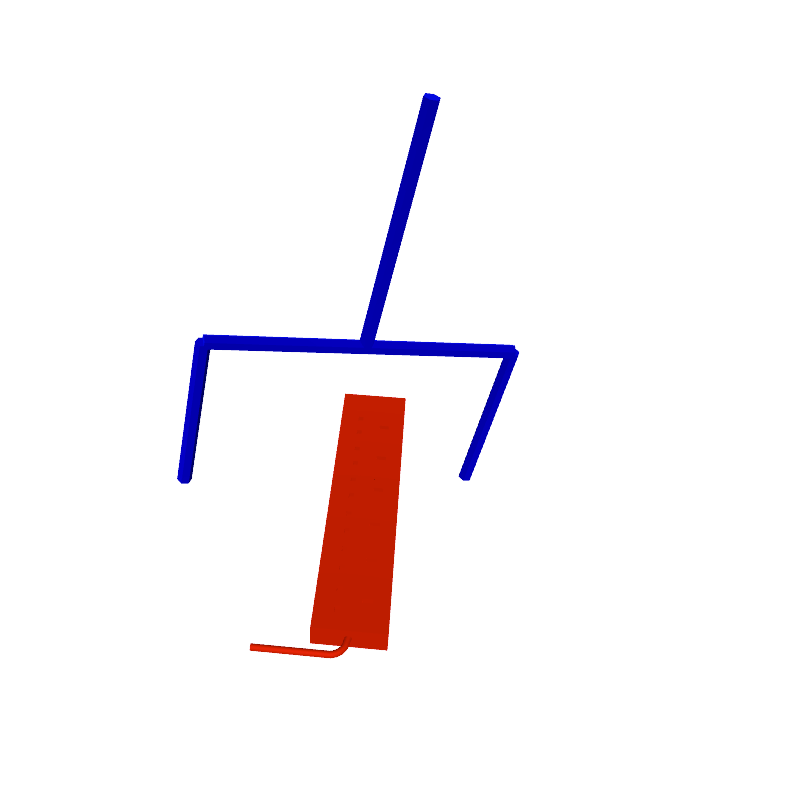}} & Flip the power strip over to inspect the information printed on its underside. \\
        \bottomrule
    \end{tabular}
    \caption{Representative grasp annotations from \ourdata{}, consisting of a grasp on an object with a corresponding task instruction. Note how different tasks require different grasps on the same object.}
    \label{fig:sample-task-grasps}
\end{figure}

In Section~\ref{sec:datagen}, we outline how generated possible tasks and grasps that would accomplish those tasks for each object, and then match those grasp descriptions with those from the ACRONYM dataset using GPT-4o. We illustrate sample matched tasks and grasps in Figure~\ref{fig:sample-task-grasps} and give the system and user prompts for this process in Figure~\ref{fig:prompt-grasp-matching}.

\subsection{TaskGrasp-Image Registration}
\label{appendix:taskgrasp-image}

In Section~\ref{sec:taskgrasp-image} we describe how the TaskGrasp-Image benchmark preserves the ground-truth annotations of TaskGrasp, while eliminating noise and artifacts by placing them in the context of real-world images. We now describe how we derive TaskGrasp-Image.

Recall that TaskGrasp fuses multiple RGB-D observations to create a segmented point cloud, which is used for grasp annotation. TaskGrasp-Image is created by transforming these annotated grasps back into each captured image frame. Unfortunately, the transformation between the point cloud and the RGB-D views were not published, but we can recover them using pointcloud registration techniques.

First, for each image we use GroundingDINO \cite{liu2023grounding} and SAM2 \cite{ravi2024sam2} to segment out the object mask. Using the depth map, we then backproject a partial 3D point cloud of the object, which we can register with the fused object point cloud. To do so, we first use DeepGMR \cite{yuan2020deepgmr} to roughly align the point clouds, and then refine using iterative closest point (ICP). Finally, we reject failed registrations by rejecting any with a final residual error exceeding 0.006. This results in 299 successfully registered views.

\begin{figure}[htb]
    \centering
    \includegraphics[width=0.5\linewidth]{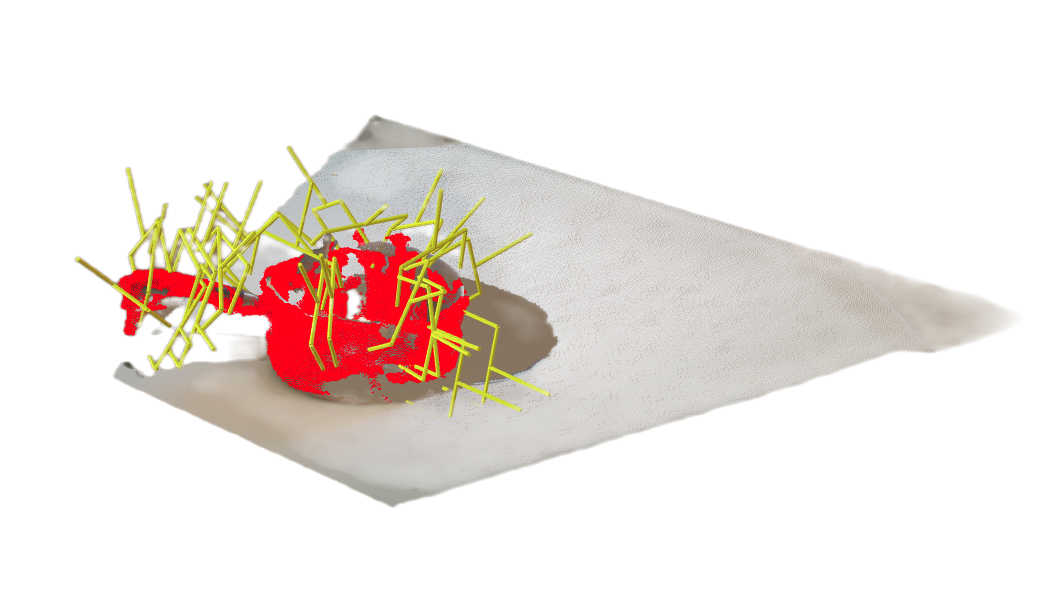}
    \caption{A visualization of the segmented pointcloud (red) of a pot registered to one of the input RGB-D views. By performing this registration, we can transform the annotated grasps (yellow) into the camera frame.}
    \label{fig:taskgrasp-image-viz}
\end{figure}

Now that we have recovered the transform between the fused point clouds and the original \mbox{RGB-D} views, we transform the annotated grasps from the point cloud into the camera frame of each viewpoint. We visualize an representative result of this process in Figure~\ref{fig:taskgrasp-image-viz}.

\subsection{Training \ours{}}
\label{appendix:graspmolmo_training_details}

Since \ours{} outputs a point on the image plane, we must supervise the training with 2D points rather than 6-DOF grasps. To map each grasp in \ourdata{} to a point on the image, we perform the following process. For each sampled grasp on each object, as described in Section~\ref{appendix:grasp_sampling}, we raycast from the grasp TCP along the grasp direction onto the object mesh. The resulting point on the object mesh is the ``grasp point'' corresponding to the grasp, which is used to supervise \ours{}'s point prediction. In a very small number of cases, this raycast may not intersect the object mesh (e.g. if the grasp is off-center of the mesh), in which case we simply select the point on the mesh closest to the grasp TCP.

In Section~\ref{sec:graspmolmo} we outline the training process for \ours{}, which largely follows from \cite{deitke2024molmopixmoopenweights}. Our training data mixture is mostly the same, with \ourtrainingdata{} and TaskGrasp-Image added, and the original data proportionally downweighted. The complete mixture is outlined in Table~\ref{tab:data-mixture}. We start with the \verb|Molmo-7B-D-0924| model and finetune on this data for 10,000 steps with a batch size per device of 8, on 64 Nvidia H100 GPUs, taking approximately 9 hours.

\begin{table}[htb]
\centering
\renewcommand{\arraystretch}{1.25}
\setlength{\tabcolsep}{0.4cm}
\begin{tabular}{cl}
    \toprule
    45\% & \ourtrainingdata{} \\ \hline
    10\% & TaskGrasp-Image (split \verb|t/0|) \\ \hline
    \multirow{4}{*}{15\%} & PixMo-AskModelAnything \\
    & PixMo-Cap \\
    & PixMo-CapQA \\
    & PixMo-Point-Explanations \\ \hline
    \multirow{2}{*}{20\%} & PixMo-Points \\
    & PixMo-Count \\ \hline
    \multirow{4}{*}{10\%} & VQA v2.0 (COCO 2014 subset), TextVQA, OK-VQA, \\
    & ChartQA, DocVQA, InfographicVQA, AI2D, A-OKVQA,
    \\ & AndroidControl, ScienceQA, TabMWP, ST-VQA, TallyQA, 
    \\ & DVQA, FigureQA, PlotQA, PixMo-Clocks \\
    \bottomrule
\end{tabular}
\smallskip
\caption{Training data mixture for \ours{}.}
\label{tab:data-mixture}
\end{table}

\subsection{\ours{} and Molmo Inference Details}
\label{appendix:prompting_details}

When performing inference with Molmo and GraspMolmo, specific phrasing of the prompt may affect performance. In both training and evaluation, we use the following prompt for GraspMolmo: ``\texttt{Point to the grasp that would accomplish the following task: <task>}''. For Molmo evaluation, we use the following prompt: ``\texttt{Point to where I should grasp to accomplish the following task: <task>}'', which we found to produce better results for Molmo than the former. In both prompt templates, \texttt{<task>} should be replaced with the relevant task command, e.g. ``Hand me the knife safely''.

As described in Section~\ref{sec:graspmolmo}, GraspMolmo (and therefore Molmo as well) outputs a point $p\in\mathbb{R}^2$ on the image, which must be mapped to a candidate grasp to be executed. Naively, one could consider back-projecting $p$ to 3D space using depth information and directly choosing the closest grasp in cartesian space. However, this is suboptimal due to (a) depth information being incomplete and therefore not necessarily available for every pixel, and (b) small errors in pixel space translating to large errors in 3D space at object boundaries, resulting in nearby points in pixel space being very far apart in cartesian space.

To address these problems, we instead transform every candidate grasp to pixel space and perform the closest-point matching there. While this does not completely solve the problem of disparate grasps being mapped to similar points, it at least does not incorrectly discard grasps on the object being pointed to, thereby increasing recall. Directly projecting every candidate grasp to pixel space is suboptimal, since the VLM outputs the point on the surface of the object to be grasped, which may not coincide with the origin of the grasp (i.e. the TCP), or any specific point in the grasp frame, so we opt for a slightly more sophisticated approach. Concretely, given the scene point cloud, for each grasp candidate we find every point within the (non-axis aligned) box bounded by the gripper fingers, and of those, select the one closest to the gripper origin. We then project this point to the image plane to get the pixel point corresponding to the grasp candidate. Finally, finding the closest such point to the VLM's output yields the predicted grasp.

\subsection{Real-Robot Evaluation}
\label{appendix:real_robot_eval_details}

In Section~\ref{sec:evaluation} we test multiple TOG models on multiple challenging semantic grasping tasks in realistic and busy scenes. We enumerate the manipulated objects and the associated tasks in Table~\ref{tab:real-tasks}. Notably, we select the tasks such that the robot must grasp the object in different ways. This ensures that we test a model's ability to understand task-level semantics (where to grasp an object for a particular task) rather than simply memorizing object-level semantics (where to grasp an object independent of task).

As noted in Section~\ref{sec:limitations}, GraspMolmo has some limitations that could be addressed in future work. In our evaluations, these limitations manifest as some common failure modes. These modes include (a) pointing to the wrong part of the correct object, (b) pointing to the wrong object, and (c) pointing to the correct location, but matching the point to the incorrect grasp. Respectively, these modes account for 62\%, 24\%, and 5\% of failures on PRISM-Real.

\begin{table*}[htb]
    \centering
    \renewcommand{\arraystretch}{1.25}
    \setlength{\tabcolsep}{0.4cm}
    \begin{tabular}{ccl}
        \toprule
        \multirow{6}{*}{Scene 1} & \multirow{2}{*}{French Press} & ``Pour coffee from the french press'' \\
                                 &  & ``Press down the knob of the plunger of the french press'' \\ \cline{2-3}
                                 & \multirow{2}{*}{Kitchen Knife} & ``Use the knife to cut fruit'' \\
                                 &  & ``Hand me the knife safely'' \\ \cline{2-3}
                                 & \multirow{2}{*}{Mug} & ``Pour the water out of the blue mug'' \\
                                 &  & ``Hang the blue mug onto a hook by the handle'' \\
        \midrule
        \multirow{6}{*}{Scene 2} & \multirow{2}{*}{Water Bottle} & ``Open the lid of the water bottle'' \\
                                 &  & ``Give me some water'' \\ \cline{2-3}
                                 & \multirow{2}{*}{Sink} & ``Adjust the faucet'' \\
                                 &  & ``Turn on the sink'' \\ \cline{2-3}
                                 & \multirow{2}{*}{Spray Bottle} & ``Spray cleaning solution with the spray bottle'' \\
                                 &  & ``Unscrew the spray bottle'' \\
        \midrule
        \multirow{6}{*}{Scene 3} & \multirow{2}{*}{Books} & ``Pass the book written by Greg Bear'' \\
                                 &  & ``Pass the book written by Orson Scott Card'' \\ \cline{2-3}
                                 & \multirow{2}{*}{Telephone} & ``Answer the phone'' \\
                                 &  & ``Put the phone back on the hook '' \\ \cline{2-3}
                                 & \multirow{2}{*}{Flower + Vase} & ``Take the flowers out of the vase'' \\
                                 &  & ``Dump the flowers out of the vase'' \\
        \bottomrule
    \end{tabular}
    \caption{We evaluate on a variety of \textbf{real-world objects in multiple scenes} representative of in-home use cases. For each object, we test multiple tasks which require different grasping affordances.}
    \label{tab:real-tasks}
\end{table*}

\subsection{Bimanual Task-Oriented Grasping Details}
\label{appendix:bimanual_details}

In Section~\ref{sec:bimanual} we provide a preliminary evaluation of GraspMolmo on a bimanual platform, allowing it to accomplish a wider variety of tasks with more complexity compared to single-arm TOG systems. This is provided as a showcase of possible capabilities, but more thorough methodology and evaluation are beyond the scope of this work. To get bimanual grasps to open the water bottle, as pictured in Figure~\ref{fig:qualitative}, we query GraspMolmo twice with the tasks ``hold the water bottle'' and ``open the water bottle''. Future work can extend and improve upon task decomposition to ameliorate bimanual TOG.

\begin{figure*}[!b]
    \centering
    \begin{center}

    \begin{tcolorbox} [top=2pt,bottom=2pt, width=\linewidth, boxrule=1pt, colback=gray!5, sharp corners]
    {\footnotesize {\fontfamily{zi4}\selectfont
    \verbatimfont{\fontfamily{zi4}}
    \begin{verbatim}
SYSTEM PROMPT:
You are an expert in robotic grasp analysis. Your task is to generate precise and
concise descriptions of robotic grasps in images. Each image contains a robotic
gripper interacting with an object. A red rectangle marks the area between the
fingers of the gripper, which helps you identify the grasp location. Your goal
is to describe where the gripper is grasping the object and what the fingers are
pinching. Follow these guidelines:

- Clearly specify the grasp location on the object (e.g., "on the handle",
    "near the rim", "on the body", "at the base").
- Indicate how the fingers of the gripper interact with the object (e.g.,
    "gripping the inner and outer surfaces," "pinching from opposite sides").  
- The image may contain multiple viewpoints, but your description should focus
    on the grasp itself rather than commenting on different perspectives.  
- Do not speculate about grasp stability or effectiveness.  
- Keep the description concise but detailed, focusing only on the grasp.
- Do not mention the red rectangle in your description, it is only for
    visualization.

Example of a good description:
"The grasp is on the rim of the pan, approximately opposite the handle. 
    The fingers are gripping the inside and outside of the pan's rim."

Your response should always describe the grasp clearly and concisely without
asking for additional input.

USER PROMPT:
These are multiple views of a(n) <object_name>. Describe the grasp.
    \end{verbatim}
    }}
    \end{tcolorbox}

    \end{center}
    \caption{System and user prompt used with GPT-4o to generate synthetic grasp descriptions for grasps in the ACRONYM dataset.}
    \label{fig:prompt-synthetic-grasp-description-generation}
\end{figure*}

\begin{figure*}[!b]
    \centering
    \begin{center}
    
    \begin{tcolorbox} [top=2pt,bottom=2pt, width=\linewidth, boxrule=1pt, colback=gray!5, sharp corners]
    {\footnotesize {\fontfamily{zi4}\selectfont
    \verbatimfont{\fontfamily{zi4}}
    \begin{verbatim}
For objects of a given category, we need to determine two grasp definitions
considering a single 6-DOF end effector that are as varied as possible (e.g.,
for ladles we would probably define one grasp around the handle and another one
around the bowl, possibly requiring different relative orientations of the
gripper with respect to the object part) and are reasonable for specific
tasks/contexts (think about different grasps when using, cleaning, or handing a
knife). Assuming that the objects from each category are standing or lying on a
table or similar surface, avoid grasps that assume that the object needs to be
approached from underneath or example tasks that require the object to be placed
upright while holding it from underneath for the target object pose on some
surface. Do not assume the presence of any optional feature in an object of the
given category (e.g. some chair subtypes might have legs, but others a wheeled
base instead, while all chairs have a back rest and a seat). Related, if the
object category is too generic to identify object parts (e.g. an undetermined
`tool`), feel free to return an empty list. Generate the output as a JSON dict
with:
 - `object_subtypes` (list of str, possibly empty) different types of object for
    the given category (which might include specific optional parts),
 - `optional_parts` (list of str, possibly empty) present in only some objects
    of the category and subtypes and should be avoided
 - `common_parts` (list of str, possibly empty) reasonably comprehensive list of
    parts of any object of the given category (no optional ones). Here you can
    merge parts that might be differently names in subclasses but offer a common
    affordance for any object of the category or subtype(s)
 - `object_table_contact_parts` (list of str) part(s) of object that are assumed
    to be in contact with the table or surface underneath in the default starting
    pose (this defines a starting object orientation). If more than one part,
    make sure these are plausibly simultaneously in contact with the underlying
    surface,
 - `common_graspable_parts` (list of str, possibly empty) that can be used to
    generate grasps for any object of the category
 - `grasps`, a list of two dicts with the entries:
   - `object_part` (str),
   - `example_task`(str),
   - `approach_direction` (str, for wrist axis, e.g. from above, from the side, 
      from below, at an angle, if relevant, else `Any`, relative to the
      orientation implied by the object-table contact parts),
   - `finger_plane` (str, relative to wrist axis, e.g. left/right or up/down
      relative to the arm's axis, if relevant, else `Any`),
   - `gripper_orientation` (str, whether the gripper faces up, down, sideways,
      or is diagonal/angled tilted, if relevant, else `Any`),
   - `natural_language` (str, description of the grasp in natural language,
      avoiding any reference to the example task and avoiding irrelevant grasp
      parameters, if any)

Make sure that example tasks do not state the object part to contact or the
direction to approach, and are unfeasible for the alternative grasp(s) in the
list. If these requirements seem impossible to fulfill, it is best to return an
empty list of grasps. Be very descriptive about the relative gripper
orientations.

One example for the `drill` category would be:
[DRILL ANNOTATION EXAMPLE]

Feel free to discuss options to annotate the object type while fulfilling the
requirements before generating the JSON dict. The object type to annotate is
[CATEGORY].
    \end{verbatim}
    }}
    \end{tcolorbox}

    \end{center}
    \caption{\textbf{Prompt template for grasp description} used in the first step of task generation.}
    \label{fig:prompt-grasp-description-generation}
\end{figure*}

\begin{figure*}[!b]
    \centering
    \begin{center}
    
    \begin{tcolorbox} [top=2pt,bottom=2pt, width=\linewidth, boxrule=1pt, colback=gray!5, sharp corners]
    {\footnotesize {\fontfamily{zi4}\selectfont
    \verbatimfont{\fontfamily{zi4}}
    \begin{verbatim}
We need to generate semantic manipulation tasks requiring each of the given
grasps in the list provided at the end. Please generate the tasks for each grasp
with the following design criteria, where each criterion is first identified by
a short name and then described in more detail:
  1. Clear target. Ensure that every task mentions the object type (e.g., 'the
     mug') unless it is obvious without it.
  2. Unknown state. Avoid tasks that make assumptions about the state of the
     object (e.g. being open/closed, empty/full, etc.).
  3. Unknown context. Avoid tasks that make assumptions about the 
     surroundings/context of the object (i.e. assuming the presence of any other
     objects of the same category or others in the scene, other than the presence
     of a table top or similar surface underneath the object at the start).
  4. Implicit grasp. Avoid references to the part of the object being grasped
     (e.g., 'by the handle') or any of the grasp definition parameters in the
     task definition.
  5. Single gripper. While you should favor single-gripper task definitions, if a
     second gripper is implied or required, it should not be assumed to be 
     present for the initial grasp, but rather during a subsequent step (e.g. if 
     'while another gripper does ...' seems reasonable, convert it into 'for 
     posterior...').
  6. Physical plausibility. Avoid tasks that require physically implausible
     configurations, like the object being placed standing on some surface while
     held from underneath.
  7. Compact instruction. Write tasks in compact and intelligible natural 
     language and avoid technical formating like snake case.
  8. Semantic meaning. Avoid simple pick and place tasks, and try to focus on
     semantic tasks, i.e., they should rely on some affordance of the object or
     consider some compositional task where we must manipulate the object towards
     some meaningful goal.
  9. Identifiability. If both provided grasps, object category or parts to grasp
     seem too coarse/vague/hard to identify, avoid defining any task and favor an
     empty list of tasks for each grasp.
   
Try to generate four valid semantic tasks per grasp, making sure that the tasks
are incompatible with the alternative grasp for the object category (they should
imply different use cases or affordances). For each generated semantic task we
need a dict with the entries:
 - `text`: the semantic task instruction, without mentioning the grasped part or
    approach direction, and mentioning the target object if needed,
 - `num_grippers`: the number of grippers required to complete the semantic task,
 - `grasp_critique`: short string justifying the lack of validity of the
    assigned grasp towards completing the task,
 - `grasp_score`: validity score in range 0 (low) to 9 (high) based on the
    grasp_critique,
 - `alternative_grasp_critique`: short string justifying the possible validity
    of the alternative grasp towards completing the task,
 - `alternative_grasp_score`: validity score in range 0 to 9 according based on
    the alternative_grasp_critique,
 - `weakest_point`: short name (string) of the task design criterion point most
    poorly fulfilled,
 - `task_criteria_fulfilled`: score the fulfillment of the weakest point in the
    range 0 (poor) to 9 (perfect fulfillment)
    
Feel free to reason about the problem and generate a JSON dictionary mapping
each grasp id to the list of semantic task dicts.

The following are the valid grasp ids and corresponding info for an object of
type '[CATEGORY]' assuming the object is in contact with the underlying surface
through its part(s) [OBJECT SURFACE CONTACTS]:
[GRASP INFOS WITHOUT EXAMPLE TASKS]
    \end{verbatim}
    }}
    \end{tcolorbox}

    \end{center}
    \caption{\textbf{Prompt template for semantic task generation,} used in the second step.}
    \label{fig:prompt-semantic-task-generation}
\end{figure*}

\begin{figure*}[!b]
    \centering
    \begin{center}

    \begin{tcolorbox} [top=2pt,bottom=2pt, width=\linewidth, boxrule=1pt, colback=gray!5, sharp corners]
    {\footnotesize {\fontfamily{zi4}\selectfont
    \verbatimfont{\fontfamily{zi4}}
    \begin{verbatim}
SYSTEM PROMPT:
You are a linguistic and robotic expert. You are tasked with matching a candidate
grasp description to one or more of multiple options, called annotated grasp
descriptions.

You will be given a candidate grasp description, which is a description of how a
robot could grasp a specific object. You will also be given a list of annotated
grasp descriptions, which are multiple known descriptions of how a robot could
grasp the same object. You should choose the annotated grasp descriptions that
have the same meaning as the candidate grasp description. In this case, "meaning"
means that the candidate grasp description and the annotated grasp description
describe a grasp on a similar part of the object, in a similar manner.

For example, if the candidate grasp description is "grasp the midpoint of the handle
of the mug", and one of the annotated grasp descriptions is "grasp the handle of the
mug", then you should choose that annotated grasp description. If there are multiple
annotated grasp descriptions that have the same meaning as the candidate grasp
description, you should return all of them. If there are no suitably matching
annotated grasp descriptions, you should return an empty list.

You should output a JSON object with the following fields:
- candidate_grasp_desc: the candidate grasp description which you are prompted with
- matching_grasp_descs: a list of annotated grasp descriptions that have the same
    meaning as the candidate grasp description

USER PROMPT:
The object is a(n) <object_category>. The candidate grasp description is:
"<candidate_grasp>". The annotated grasp descriptions are:
- <grasp_description_1>
- <grasp_description_2>
- ...
    \end{verbatim}
    }}
    \end{tcolorbox}

    \end{center}
    \caption{System and user prompt used with GPT-4o to match generated candidate grasp descriptions with synthetically generated descriptions of ACRONYM grasps.}
    \label{fig:prompt-grasp-matching}
\end{figure*}

\end{document}